\documentclass[10pt,twocolumn,letterpaper]{article}

\usepackage{iccv}
\usepackage{times}
\usepackage{epsfig}
\usepackage{graphicx}
\usepackage{amsmath}
\usepackage{amssymb}
\usepackage{booktabs}
\usepackage{float} 
\usepackage{algorithm}
\usepackage{algorithmic}
\usepackage{multirow}
\usepackage{wrapfig}
\usepackage[table]{xcolor}
\usepackage{pifont}
\usepackage{subcaption}
\usepackage[accsupp]{axessibility}  

\usepackage[pagebackref=true,breaklinks=true,letterpaper=true,colorlinks,bookmarks=false]{hyperref}

\newcommand{\collapser}[1]{}
\newcommand{\ourmethod}{SSB\xspace}
\newcommand{\mathbbm}[1]{\text{\usefont{U}{bbm}{m}{n}#1}}
\newcommand{\myparagraph}[1]{\vspace{2pt}\noindent{\bf #1}}

\newcommand{\camrdy}[1]{\textcolor[rgb]{0.0, 0.0, 0.0}{#1}}
\newcommand{\none}[1]{\textcolor[rgb]{1.0, 1.0, 1.0}{#1}}

\iccvfinalcopy 

\def\iccvPaperID{5179} 

\ificcvfinal\fi

\begin{document}

\title{
SSB: Simple but Strong Baseline for Boosting Performance of\\ Open-Set Semi-Supervised Learning
}

\author{Yue Fan \qquad Anna Kukleva \qquad Dengxin Dai \qquad Bernt Schiele\\
{\tt\small \{yfan, akukleva, ddai, schiele\}@mpi-inf.mpg.de 
}\\
Max Planck Institute for Informatics, Saarbrücken, Germany
\\ Saarland Informatics Campus
}

\maketitle
\ificcvfinal\thispagestyle{empty}\fi

\begin{abstract}
Semi-supervised learning (SSL) methods effectively leverage unlabeled data to improve model generalization. 
However, SSL models often underperform in open-set scenarios, where unlabeled data contain outliers from novel categories that do not appear in the labeled set.
In this paper, we study the challenging and realistic open-set SSL setting, where the goal is to both correctly classify inliers and to detect outliers.
Intuitively, the inlier classifier should be trained on inlier data only. 
However, we find that inlier classification performance can be largely improved by incorporating high-confidence pseudo-labeled data, regardless of whether they are inliers or outliers. 
Also, we propose to utilize non-linear transformations to separate the features used for inlier classification and outlier detection in the multi-task learning framework, preventing adverse effects between them.
Additionally, we introduce pseudo-negative mining, which further boosts outlier detection performance. 
The three ingredients lead to what we call \textbf{S}imple but \textbf{S}trong \textbf{B}aseline (\ourmethod) for open-set SSL. 
In experiments, \ourmethod greatly improves both inlier classification and outlier detection performance, outperforming existing methods by a large margin. 
Our code will be released at \url{https://github.com/YUE-FAN/SSB}.

\end{abstract}

\section{Introduction} \label{sec:intro}
\begin{figure}[t]
\centering
\begin{minipage}{0.49\linewidth}
    \centering
    \includegraphics[width=\linewidth]{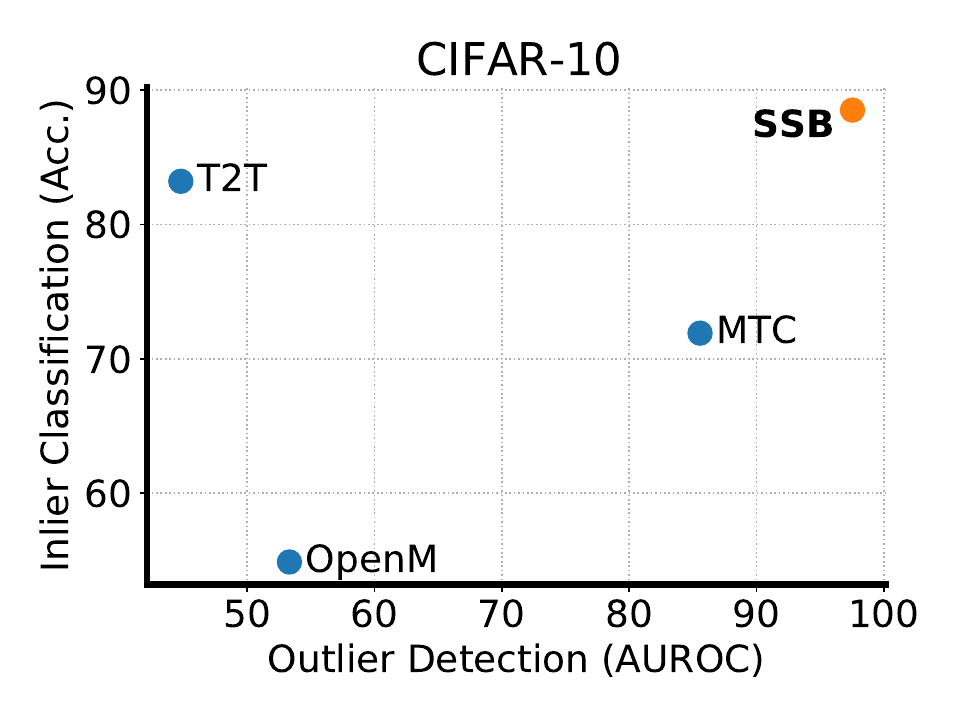}
\end{minipage}%
\begin{minipage}{0.49\linewidth}
    \centering
    \includegraphics[width=\linewidth]{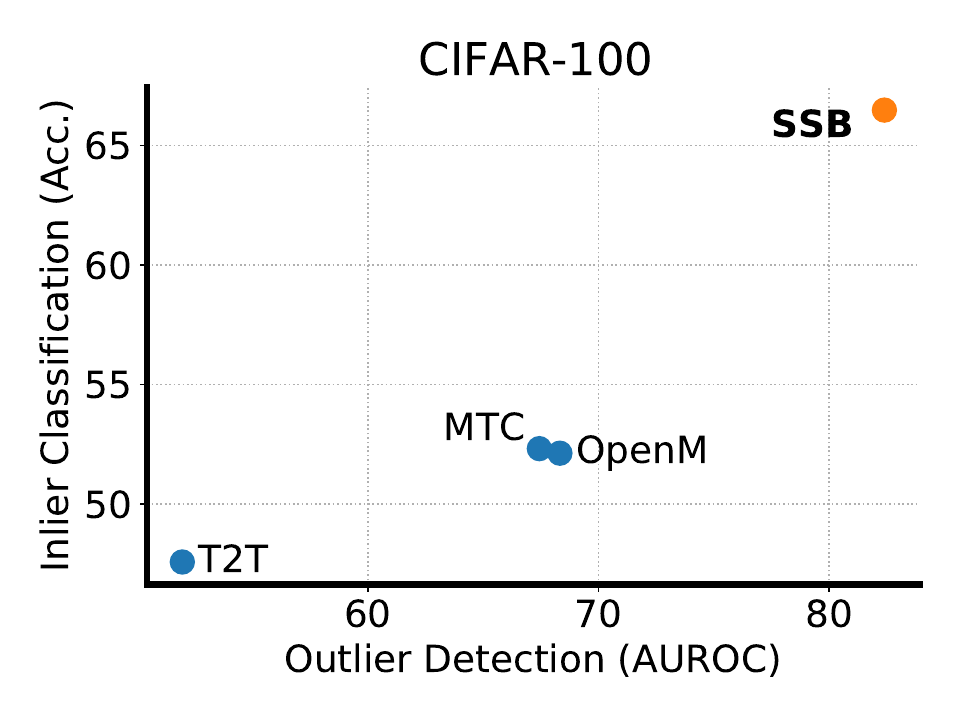}
\end{minipage}
\begin{minipage}{\linewidth}
    \centering
    \includegraphics[width=\linewidth]{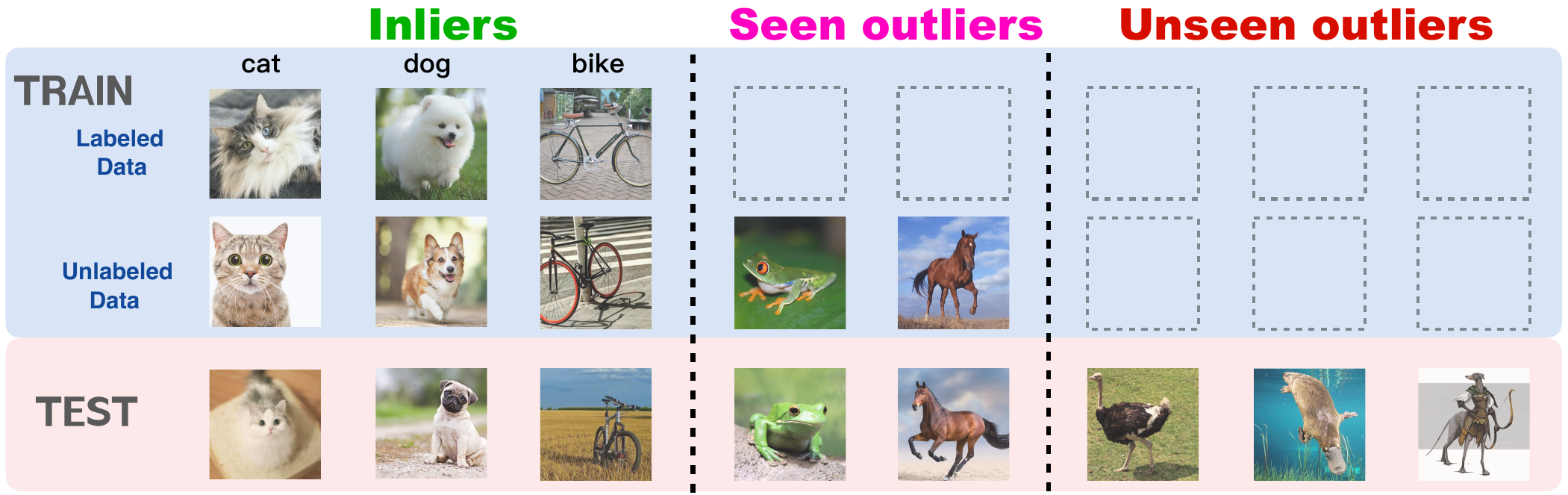}
\end{minipage}
\caption{
Open-set semi-supervised learning considers a realistic and challenging setting, where unlabeled data contains samples from novel classes (\textcolor[rgb]{1, 0, 0.756}{\textbf{seen outliers}}) that do not appear in the labeled data.
At test time, the model should correctly classify \textcolor[rgb]{0, 0.694, 0}{\textbf{inliers}}, while identifying outliers seen during the training and, most importantly, \textcolor[rgb]{0.843, 0.050, 0}{\textbf{unseen outliers}} that do not appear in the training set.
We measure test accuracy for the inlier classification performance and AUROC for the outlier detection performance.
Our method (\ourmethod) achieves superior performance in both tasks. 
}
\label{fig:teaser_setting}
\end{figure}

Semi-supervised learning (SSL) has achieved great success in improving model performance by leveraging unlabeled data \cite{lee2013pseudo,laine2016pimodel,tarvainen2017meanteachers,miyato2018vat,berthelot2019mixmatch,berthelot2019remixmatch,sohn2020fixmatch,xie2019uda,fan2021revisiting,zhang2021flexmatch,usb2022}.
However, standard SSL assumes that the unlabeled samples come from the same set of categories as the labeled samples, which makes them struggle in open-set settings \cite{oliver2018realistic}, where unlabeled data contain out-of-distribution (OOD) samples from novel classes that do not appear in the labeled set (see Fig. \ref{fig:teaser_setting}). 
In this paper, we study this more realistic setting called \textit{open-set semi-supervised learning}, where the goal is to learn both a good closed-set classifier to classify inliers and to detect outliers as shown in Fig.~\ref{fig:teaser_setting}.

Recent works on open-set SSL \cite{huang2021t2t,chen2020uasd,saito2021openmatch,yu2020mtc,guo2020ds3l,he2022safe,he2022not,huang2022tmm} have achieved strong performance 
\cite{wang2019progress,maimon2005data,hodge2004survey,aggarwal2001outlier} through a multi-task learning framework, which consists of an inlier classifier, an outlier detector, and a shared feature encoder, as shown in Figure \ref{fig:pipeline}. 
The outlier detector is trained to filter out OOD data from the unlabeled data so that the classifier is only trained on inliers. However, this framework has two major drawbacks. 
First, detector-based filtering often removes many inliers along with OOD data, leading to suboptimal classification performance due to the low utilization ratio of unlabeled data. 
Second, the inlier classifier which shares the same feature encoder with the outlier detector can have an adverse effect on the detection performance as shown in Table \ref{tab:disentangle}.

To this end, we contribute a \textbf{S}imple but \textbf{S}trong \textbf{B}aseline, \textbf{\ourmethod}, for open-set SSL with three ingredients to address the above issues.
(1) In contrast to detector-based filtering aiming to remove OOD data, we propose to incorporate pseudo-labels with high inlier classifier confidence into the training, \textit{irrespective of whether a sample is an inlier or OOD}. This not only effectively improves the unlabeled data utilization ratio but also includes many useful OOD data that can be seen as natural data augmentations of inliers (see Fig. \ref{fig:ablation_ood}).
(2) Instead of directly sharing features between the classifier and detector, we add non-linear transformations for the task-specific heads and find that this 
effectively reduces mutual interference between them, resulting in more specialized features and improved performance for both tasks.
(3) In addition, we propose pseudo-negative mining to further improve  outlier detector training by enhancing the data diversity of OOD data with pseudo-outliers.
Despite its simplicity, \ourmethod achieves significant improvements in both inlier classification and OOD detection. 
As shown in Fig. \ref{fig:teaser_setting}, existing methods either struggle in detecting outliers or have difficulties with inlier classification while \ourmethod obtains good performance for both tasks. 




\section{Related Work}
\myparagraph{Semi-supervised learning.}
Semi-supervised learning (SSL) aims to improve model performance by exploiting both labeled and unlabeled data.
As one of the most widely used techniques, pseudo-labeling \cite{lee2013pseudo} is adopted by many strong SSL methods \cite{sohn2020fixmatch,berthelot2019mixmatch,berthelot2019mixmatch,xie2019uda,zhang2021flexmatch,arazo2020plcb,pham2020meta,xie2020noisestudent,berthelot2021adamatch,li2021comatch}.
The idea is to generate artificial labels for unlabeled data to improve  model training.
\cite{berthelot2019mixmatch,berthelot2019remixmatch} compute soft pseudo-labels and then apply MixUp \cite{zhang2017mixup} with labeled data to improve the performance;
\cite{sohn2020fixmatch,xie2019uda,zhang2021flexmatch} achieves good performance by combining pseudo-labeling with consistency regularization \cite{laine2016pimodel,miyato2018vat,tarvainen2017meanteachers};
\cite{pham2020meta} proposes a meta learning approach that uses a teacher model to refine the pseudo-labels based on the training of a student model;
\cite{xie2020noisestudent} leverages the idea of self-training which generates pseudo labels in an iterative way and inject noise to each training stage.
In this paper, we also adopt a simple confidence-based pseudo-labeling \cite{sohn2020fixmatch} for classifier training, which is an effective way of leveraging unlabeled data to improve the model performance.
Compared to standard SSL, \ourmethod has an additional outlier detector, which enables the model to reject samples that do not belong to any of the inlier classes. 


\myparagraph{Open-set SSL \& Class-mismatched SSL.} 
First shown by \cite{oliver2018realistic}, standard SSL methods suffer from performance degradation when there are out-of-distribution (OOD) samples in unlabeled data.
Since then, various approaches have been proposed to address this challenge \cite{chen2020uasd,guo2020ds3l,yu2020mtc,saito2021openmatch,huang2021t2t,park2022opencos,he2022safe,he2022not,huang2022tmm}.
Existing methods seek to alleviate the effect of OOD data by filtering them out in different ways so that the classification model is trained with inliers only.
For example, \cite{chen2020uasd} uses model ensemble \cite{schapire1990ensemble} to compute soft pseudo-labels and performs filtering with a confidence threshold;
\cite{guo2020ds3l} proposes a bi-level optimization to weaken the loss weights for OOD data;
\cite{yu2020mtc} assigns an OOD score to each unlabeled data and refines it during the training;
\cite{saito2021openmatch} leverages one-vs-all (OVA) classifiers \cite{saito2021ovanet} for OOD detection and propose a consistency loss to train them;
\cite{huang2021t2t} proposes a cross-modal matching module to detector outliers.
\cite{huang2022tmm} employs adversarial domain adaptation to filter unlabeled data and find recyclable OOD data to improve the performance;
\cite{he2022safe} uses energy-discrepancy to identify inliers and outliers. 
In contrast, we show that if the representations of the inlier classifier and the outlier detector are well-separated, OOD data turns out to be a powerful source to improve the inlier classification without degrading the detection performance.
So, instead of filtering OOD data, we use a simple confidence-based pseudo-labeling to incorporate them into the training.




\myparagraph{Open-world SSL.} Open-set SSL is similar to open-world SSL \cite{cao2021open,rizve2022openldn,rizve2022towards} but bears several important differences.
While both have unlabeled data of novel classes during the training, the goal of open-world SSL is to classify inliers and discover new classes from OOD data instead of rejecting them.
Another important difference is that open-world SSL is often a transductive learning setting while open-set SSL requires generalization beyond the current distribution.
Namely, the model should be able to detect OOD data from novel classes that present in the training set as well as OOD data from classes that are never seen during  training.





\section{\ourmethod: Simple but Strong Baseline for Open-Set Semi-Supervised Learning} \label{sec:method}

In this section, we first present the problem setup of open-set semi-supervised learning (SSL). 
Then, we give an overview of our method \ourmethod in Section \ref{sec:method_overview} before presenting details of the three simple yet effective ingredients used in our method in Section \ref{sec:method_clf}, \ref{sec:method_nonlinear}, and \ref{sec:method_det}.

\begin{figure*}[ht]
\centering
\includegraphics[width=0.98\textwidth]{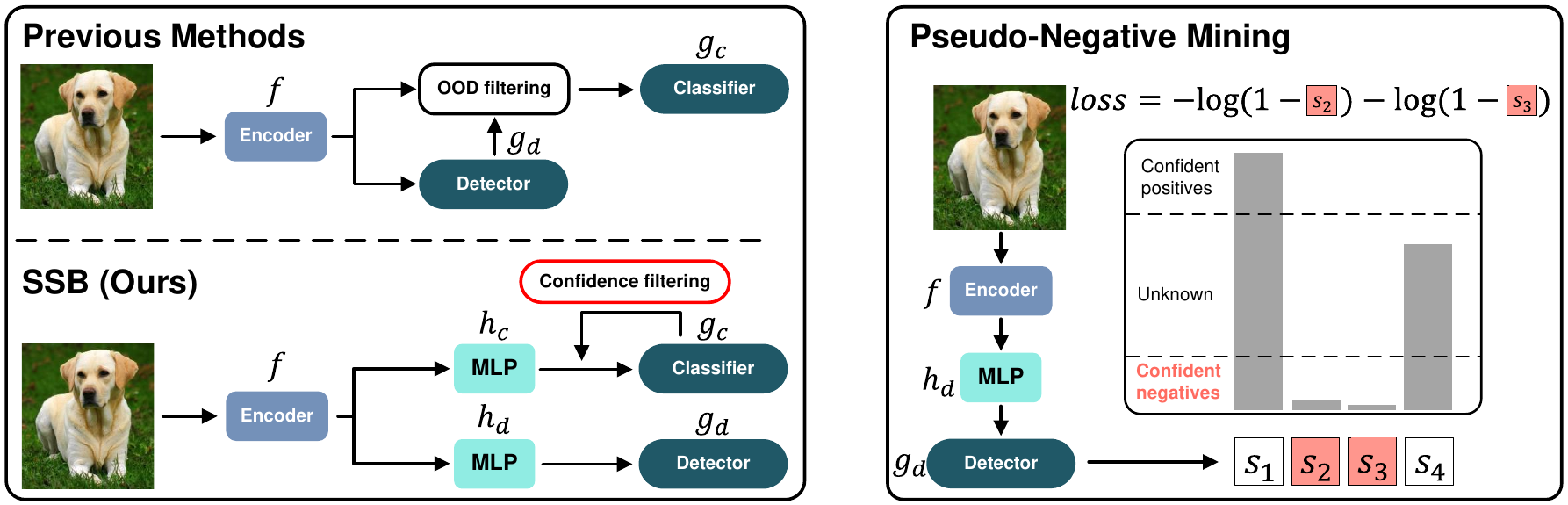}
\caption{
\textbf{Left:} 
Our baseline for open-set SSL consists of an inlier classifier $g_c$, an outlier detector $g_d$, and a shared feature encoder $f$ whose features are separated from the task-specific heads by two projection heads $h_c$ and $h_d$.
Unlike the detector-based filtering, we adopt confidence-based pseudo-labeling by the inlier classifier to leverage useful OOD data for classifier training.
For detector training, we train one-vs-all (OVA) classifiers as in OpenMatch~\cite{saito2021openmatch}.
\textbf{Right:}
Given the inlier scores ($s_1$ to $s_4$), pseudo-negative mining selects confident negatives ($s_2$ and $s_3$ in the figure), whose inlier scores are lower than a pre-defined threshold, as pseudo-outliers to help the outlier detector training.
}
\label{fig:pipeline}
\end{figure*}

\myparagraph{Problem setup and notations:}
As shown in Fig. \ref{fig:teaser_setting}, open-set SSL generalizes the settings of standard SSL and out-of-distribution (OOD) detection. 
It considers three disjoint sets of classes: $\mathcal{C}$ corresponds to the inlier classes that are partially annotated, $\mathcal{U_S}$ contains the outlier classes seen during training but without annotations, and lastly, $\mathcal{U_U}$ is composed of the classes that are not seen during training (only seen at test time).
The training data contains a small labeled set $\mathcal{D}_{\text{labeled}}=\{(\textbf{x}_i^l, y_i)\}_{i=1}^N \subset \mathcal{X} \times \mathcal{C}$ and a large unlabeled set $\mathcal{D}_{\text{unlabeled}}=\{(\textbf{x}_i^u)\}_{i=1}^M \subset \mathcal{X}$, where $\mathcal{X}$ is the input space. 
While the labeled set only consists of samples of inlier classes, 
the unlabeled set contains both samples from $\mathcal{C}$ and 
$\mathcal{U_S}$. Thus, the 
the ground-truth label of $\textbf{x}^u$ is from $\mathcal{C} \cup \mathcal{U_S}$ with $\mathcal{C} \cap \mathcal{U_S} = \emptyset$. 

The goal of open-set SSL is to train a model that can perform good inlier classification as well as detecting both seen and unseen outliers. 
Without loss of generality, consider a test set $\mathcal{D}_{\text{test}}=\{(\textbf{x}_i, y_i)\}_{i=1}^N \subset \mathcal{X} \times (\mathcal{C} \cup \mathcal{U_S} \cup \mathcal{U_U})$, where $\mathcal{C} \cap \mathcal{U_U} = \emptyset$ and  $\mathcal{U_S} \cap \mathcal{U_U} = \emptyset$. 
The learned model should be able to correctly classify inliers $\{(\textbf{x}_i| y_i \in \mathcal{C})\}$ and detect outliers from $\{(\textbf{x}_i| y_i \in \mathcal{U_S})\}$ as well as $\{(\textbf{x}_i| y_i \in \mathcal{U_U})\}$, which is crucial for practical applications. 

\subsection{Method Overview} \label{sec:method_overview}
Following \cite{huang2021t2t,chen2020uasd,saito2021openmatch,yu2020mtc,guo2020ds3l,he2022safe,he2022not,huang2022tmm}, we adopt a multi-task learning framework for open-set SSL, which performs inlier classification and outlier detection. 
As shown in Fig. \ref{fig:pipeline}, \ourmethod comprises four components: (1) An inlier classifier $g_c$, (2) an outlier detector $g_d$, (3) a shared feature encoder $f$, and (4)  importantly, two projection heads $h_c$ and $h_d$. 
Inspired by \cite{saito2021openmatch}, the outlier detector $g_d$ consists of $|\mathcal{C}|$ one-vs-all (OVA) binary classifiers, each of which is trained to distinguish inliers from outliers for each single class.
Given a batch of labeled data $\textbf{X}^l = \{(\textbf{x}_i^l, y_i)\}_{i=1}^{B_l}$ and unlabeled data $\textbf{X}^u = \{(\textbf{x}_i^u)\}_{i=1}^{B_u}$, the total loss for training the model is:
\begin{equation}
    L_{total} = L_{cls}(\textbf{X}^l,\textbf{X}^u; f, h_c, g_c) + L_{det}(\textbf{X}^l, \textbf{X}^u; f, h_d, g_d)
\end{equation}
where $L_{cls}$ and $L_{det}$ are the classification and detection losses, respectively. 
For the sake of brevity, we will drop the dependencies of the loss function on $f$, $h_c$, $g_c$, $h_d$, and $g_d$ in the following.
The complete algorithm of \ourmethod is summarized by Alg. 1 in Appendix D.

During inference, the test image is first fed to the inlier classifier to compute the class prediction. 
Then, the corresponding detector is used to decide whether it is an inlier of the predicted class or an outlier. 
We explain the details of \ourmethod in the following three sections.

\subsection{Boosting Inlier Classification with Classifier Pseudo-Labeling} \label{sec:method_clf}
Existing methods for open-set SSL \cite{huang2021t2t,chen2020uasd,saito2021openmatch,yu2020mtc,guo2020ds3l} aim to eliminate OOD data from the classifier training.
This is typically accomplished by training outlier detectors that can filter out OOD data from unlabeled data, as shown in Fig. \ref{fig:pipeline}.
However, as we will see in Table \ref{tab:clf_train}, detector-based filtering often removes many inliers along with OOD data, which leads to a low utilization ratio of unlabeled data and hinders inlier classification performance. 

In this work, instead of using detector-based filtering, we propose to incorporate unlabeled data with confident pseudo-labels (as generated by the \textit{inlier classifier}) into the training, \textit{irrespective of whether it is inlier or OOD data}. 
This not only effectively improves the unlabeled data utilization ratio but also includes many useful OOD data as natural data augmentations of inliers into the training (see Fig. \ref{fig:ablation_ood}).
Inspired by \cite{sohn2020fixmatch}, 
we train the model with pseudo-labels from the inlier classifier whose confidence scores
are above a pre-defined threshold.
Specifically, for each unlabeled sample $\textbf{x}_i^u$, we first predict the pseudo-label distribution as $\hat{p}_i^u = \text{softmax}(h_c(g_{c}(f(\textbf{x}_i^u)))$.
Then, the confidence score of the pseudo-label is computed as $\max \hat{p}_i^u$.
Finally, the cross-entropy loss is calculated for samples whose pseudo-labels have confidence scores greater than a pre-defined threshold $\tau$ as:
\begin{equation}
    L_{cls}^u(\textbf{X}^u) = \frac{1}{B_u} \sum_{i=1}^{B_u} \mathbbm{1}(\max \hat{p}_i^u \geq \tau) H(\hat{p}_i^u, \hat{y}^u_i)
\end{equation}
where $H(\cdot, \cdot)$ denotes the cross-entropy, $\hat{y}^u_i = \text{argmax} \hat{p}_i^u$, and $\mathbbm{1}(\cdot)$ is the indicator function which outputs 1 when the confidence score is above the threshold $\tau$.

The total classification loss is computed as the summation of a labeled data loss and the unlabeled data loss as:
\begin{equation}
    L_{cls}(\textbf{X}^l,\textbf{X}^u) = L_{cls}^l(\textbf{X}^l) + L_{cls}^u(\textbf{X}^u)
\end{equation}
where $L_{cls}^l$ is a standard cross-entropy loss for labeled data.

Despite its simplicity, we obtain a substantial performance improvement in inlier classification through classifier confidence-based pseudo-labeling as shown in Table \ref{tab:disentangle}.
Our method is conceptually different from previous methods as we aim to leverage OOD data rather than remove them.
On the one hand, our method effectively improves the unlabeled data utilization ratio as shown in Table \ref{tab:clf_train}, which leads to great inlier classification performance improvement.
On the other hand, our method provides an effective way of leveraging useful OOD data for classifier training.
In fact, many OOD data are natural data augmentations of inliers and are beneficial for classification performance if used carefully.
As shown in Fig. \ref{fig:ablation_ood}, the selected OOD data present large visual similarities with samples of inlier classes, and, thus, significantly enhance the data diversity, leading to improved generalization performance.

\subsection{Non-Linear Feature Boosting} \label{sec:method_nonlinear}
In previous methods, simply including OOD samples into the classifier training harms detection performance since the
inlier classifier and the outlier detector use the same
feature representation \cite{saito2021openmatch,yu2020mtc,huang2021t2t}.
On the one hand, the classifier uses OOD data as pseudo-inliers, thus mixing their representations in the feature space. 
On the other hand, the outlier detector is trained to distinguish inliers and outliers, which leads to separated representations in the feature space.
As a result, the contradiction between the classifier and the outlier detector ultimately adversely affects each other, which limits the overall performance, as shown in Table \ref{tab:disentangle}.

In this work, we find empirically that simply adding non-linear transformations between the task-specific heads and the shared feature encoder can effectively 
mitigate the adverse effect.
Given a sample $\textbf{x}_i$, two multi-layer perceptron (MLP) projection heads $h_c$ and $h_d$ are used to transform the features from the encoder.
The output of the network is thus $h_c(g_c(f(\textbf{x}_i)))$ for the classifier and $h_d(g_d(f(\textbf{x}_i)))$ for the outlier detector.
Compared to the previous methods, the non-linear transformations effectively prevent mutual interference between the classifier and detector, resulting in more specialized features and improved performance in both tasks. 
In Table \ref{tab:disentangle}, while the OOD detection performance degenerates when adding OOD data for classifier training for the model without the projection heads, \ourmethod, in contrast, still exhibits excellent performance in detecting outliers with the help of the projection heads.
Moreover, the efficacy of the non-linear projection head also generalizes to other frameworks. We show in the experiment section that it is compatible with various SSL backbones and open-set SSL methods and leads to performance improvement.

\subsection{Outlier Detection with Pseudo-Negative Mining} \label{sec:method_det}
In this section, we first describe the outlier detector used in \ourmethod and then introduce a simple yet effective technique called pseudo-negative mining to improve the outlier detector training.

Following \cite{saito2021openmatch}, we adopt $|\mathcal{C}|$ one-vs-all (OVA) binary classifiers for OOD detection, where each OVA classifier is trained to distinguish between inliers and outliers for each individual inlier class.
Given a labeled sample $\textbf{x}_i^l$ from class $y_i$, it is regarded as an inlier for class $y_i$ and an outlier for class $k, k \neq y_i$.
Therefore, the OVA classifiers can be trained using binary cross-entropy loss on the positive-negative pairs constructed from the labeled set as:
\begin{equation}
    L_{det}^l(\textbf{X}^l) = - \frac{1}{B_l} \sum_{i=1}^{B_l} log(p_{y_i}(\textbf{x}_i^l)) + \frac{1}{K} \sum_{k\neq y_i} log(1 - p_{k}(\textbf{x}_i^l))
\end{equation}
where $p_{k}(\textbf{x}_i^l)$ is the inlier score of $\textbf{x}_i^l$ for class $k$ computed by the $k$-th detector and $K = |\mathcal{C}| - 1$.

However, due to data scarcity, it is difficult to learn good representations for outliers with labeled data only. 
To this end, we propose pseudo-negative mining to further improve the outlier detector training by leveraging confident negatives as pseudo-outliers to enhance the data diversity of OOD data.
As shown in Fig. \ref{fig:pipeline}, given an unlabeled sample $\textbf{x}_i^u$, we consider it as a pseudo-outlier for class $k$ if the inlier score for class $k$ is lower than a pre-defined threshold. 
Then, $\textbf{x}_i^u$ is used as a negative sample to calculate the cross-entropy loss of class $k$. \camrdy{The final loss for $\textbf{x}_i^u$ is the summation over all classes using it as the negative sample:}
\begin{equation}
    L_{det}^u(\textbf{x}_i^u) = - \frac{1}{\sum_k{\mathbbm{1}(p_k < \theta)}} \sum_{k=1}^{|\mathcal{C}|} \mathbbm{1}(p_k < \theta) log(1 - p_{k}(\textbf{x}_i^u))
\end{equation}
where $p_{k}$ is the inlier score from the $k$-th detector and $\mathbbm{1}(\cdot)$ is the indicator function which outputs 1 when the confidence score is less than the threshold $\theta$.
This increases the data diversity of outliers and improves generalization performance as shown in Table \ref{tab:ablation_pseudo_neg}.
Compared to standard pseudo-labels, pseudo-outliers have much higher precision because we specify which classes the sample does not belong to rather than which class it belongs to.
The latter is a more difficult task than the former.
Therefore, pseudo-negative mining is less susceptible to inaccurate predictions while increasing data utilization.

Our final loss for detector training also includes Open-set Consistency (OC) loss \cite{saito2021openmatch} and entropy minimization (EM) \cite{grandvalet2005minentropy} because they can lead to further improvement.
The overall loss for training the detector is as follows:
\begin{multline}
     L_{det}(\textbf{X}^l,\textbf{X}^u) = L_{det}^l(\textbf{X}^l) + \lambda_{det}^u L_{det}^u(\textbf{X}^u) \\
     + \lambda_{OC}^u L_{OC}^{u}(\textbf{X}^u) + \lambda_{em}^u L_{em}^{u}(\textbf{X}^u)
\end{multline}
where $\lambda_{det}^u$, $\lambda_{OC}^u$, and $\lambda_{em}^u$ are loss weights; $L_{OC}^{u}$ is the soft open-set consistency regularization loss, which enhances the smoothness of the OVA classifier with respect to input transformations; $L_{em}^{u}$ is the entropy minimization loss, which encourages more confident predictions.

\section{Experiments}
In this section, we first compare \ourmethod with existing methods in Section \ref{sec:main_results}, and then provide an ablation study and further analysis in Section \ref{sec:ablation}.

\begin{figure*}[ht]
\centering
\begin{minipage}{0.33\textwidth}
    \centering
    \includegraphics[width=\textwidth]{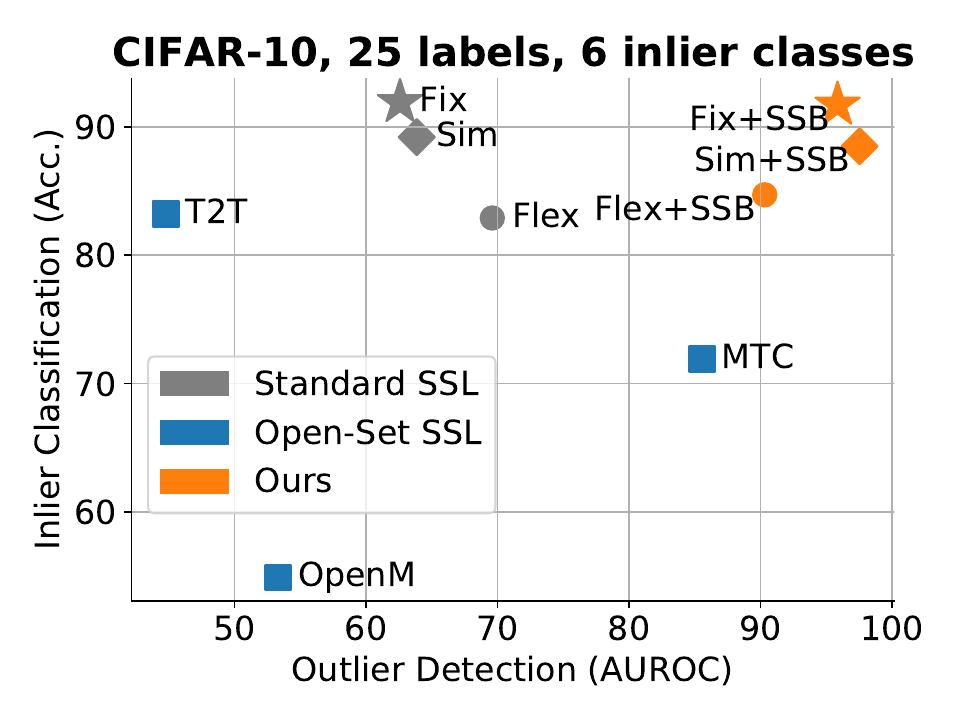}
\end{minipage}%
\begin{minipage}{0.33\textwidth}
    \centering
    \includegraphics[width=\textwidth]{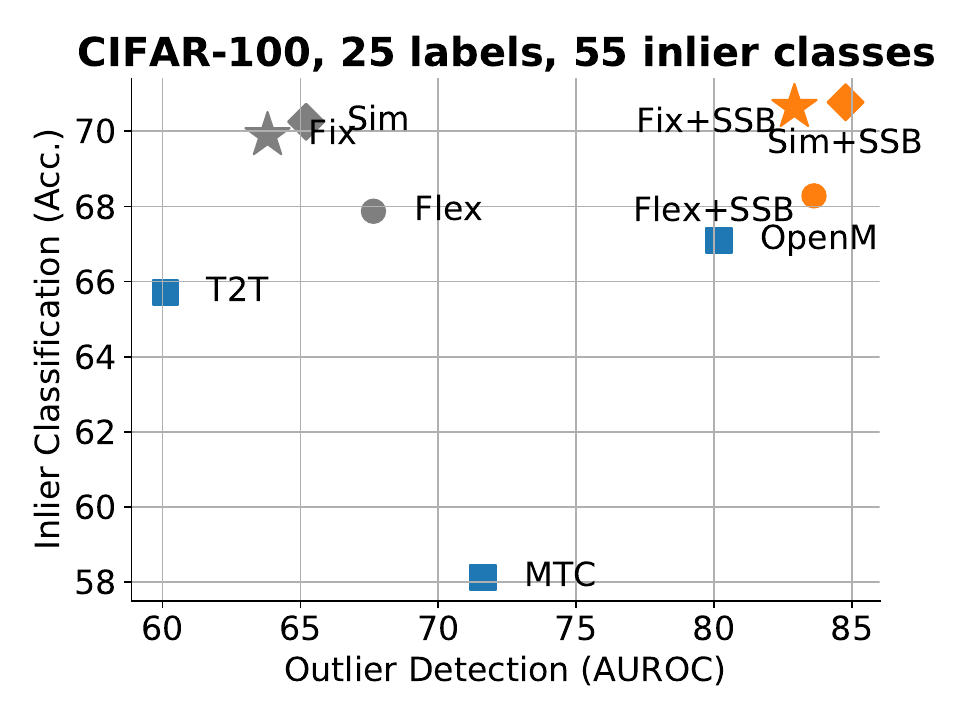}
\end{minipage}%
\begin{minipage}{0.33\textwidth}
    \centering
    \includegraphics[width=\textwidth]{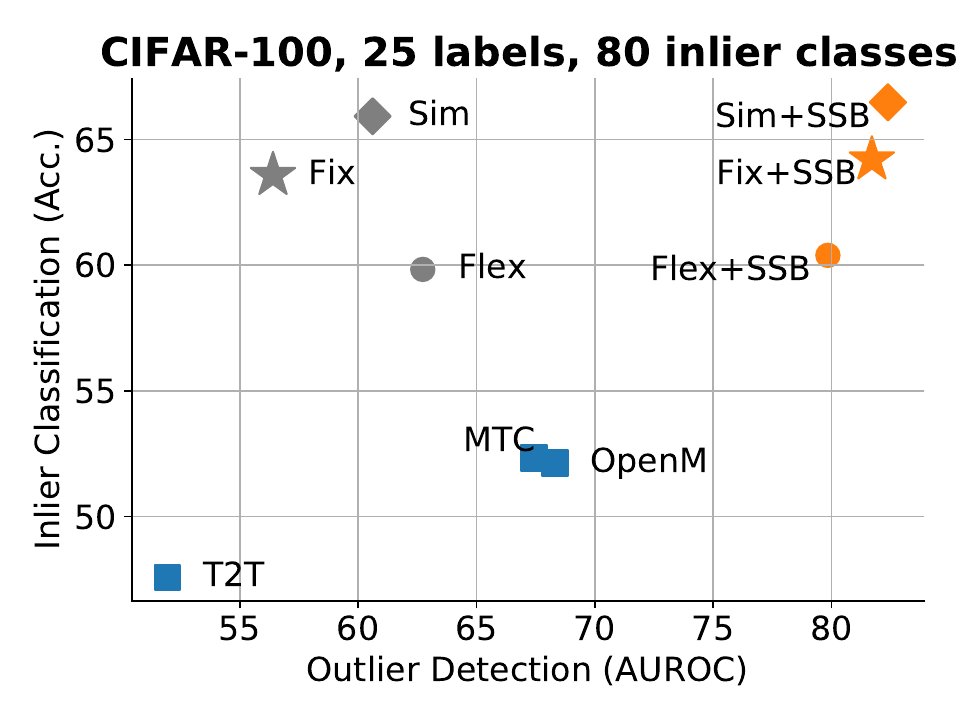}
\end{minipage}
\begin{minipage}{0.33\textwidth}
    \centering
    \includegraphics[width=\textwidth]{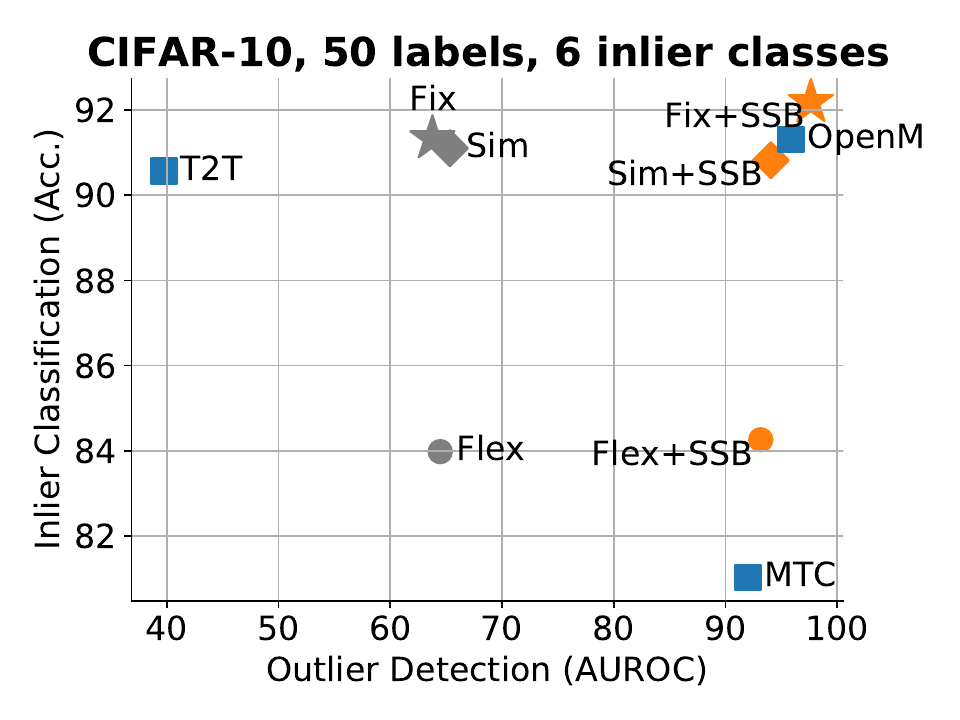}
\end{minipage}%
\begin{minipage}{0.33\textwidth}
    \centering
    \includegraphics[width=\textwidth]{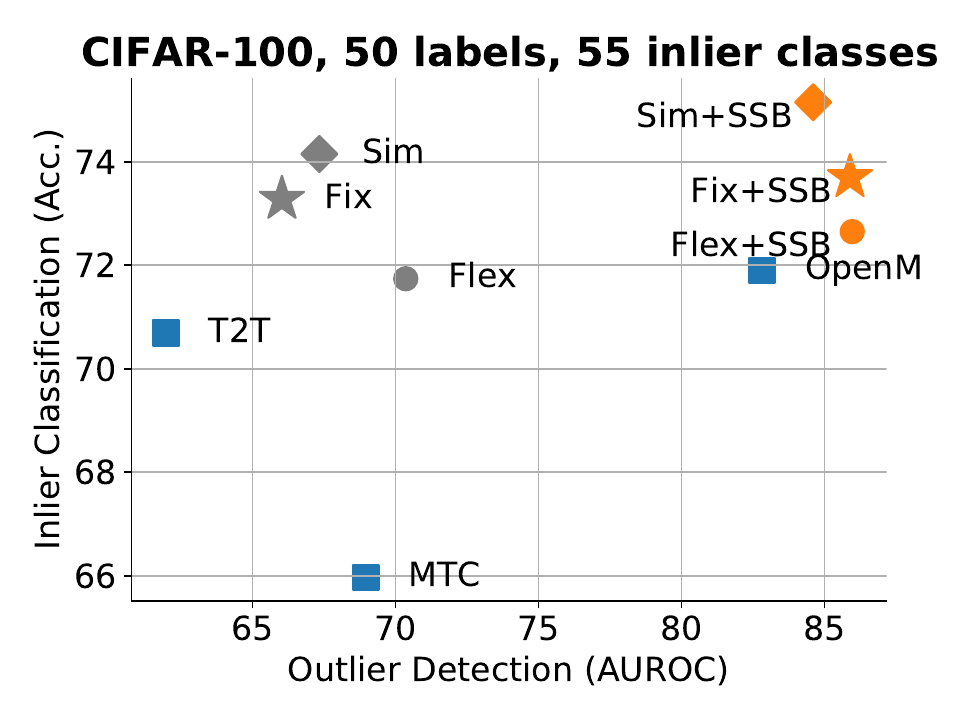}
\end{minipage}%
\begin{minipage}{0.33\textwidth}
    \centering
    \includegraphics[width=\textwidth]{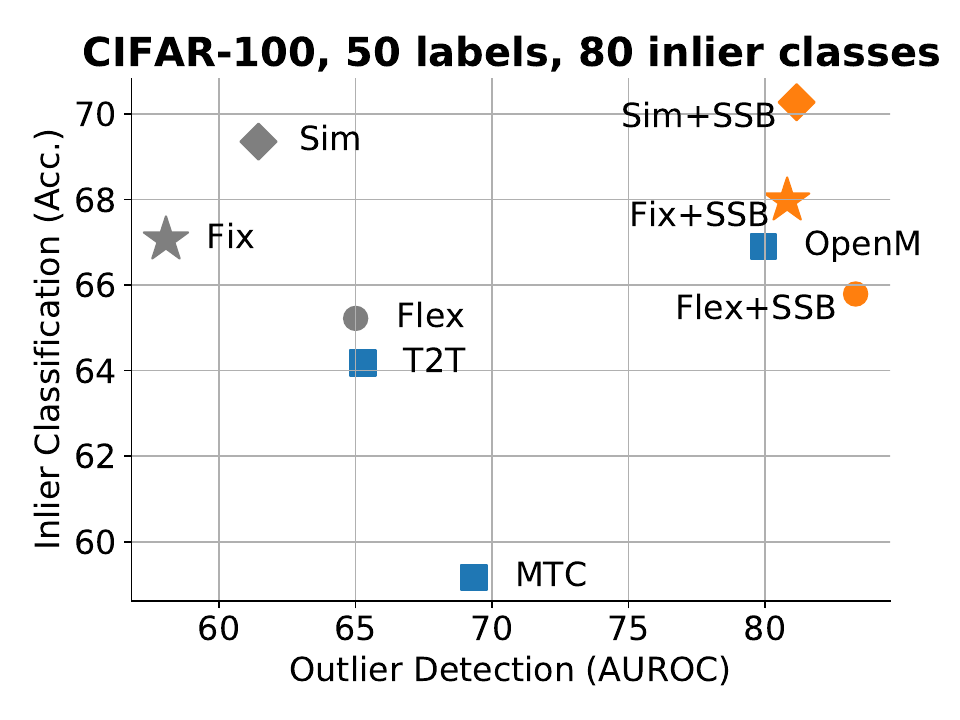}
\end{minipage}
\caption{
\textbf{Classification and detection performance on CIFAR-10 and CIFAR-100 with varying numbers of inlier classes and labeled data.}
We measure test accuracy for the inliers classification performance and AUROC for the outlier detection performance.
While standard SSL methods suffer in outlier detection and open-set SSL methods suffer in inlier classification, \ourmethod achieves good performance in both tasks. 
Noted that the reported outlier detection performance is the \textit{average AUROC in detecting both seen and unseen outliers}.
Please see Appendix A for a detailed breakdown of the results in tables and results on more benchmarks.
}
\vspace{-10pt}
\label{fig:main_results_cifar}
\end{figure*}

\begin{figure}[ht]
\centering
\includegraphics[width=0.7\linewidth]{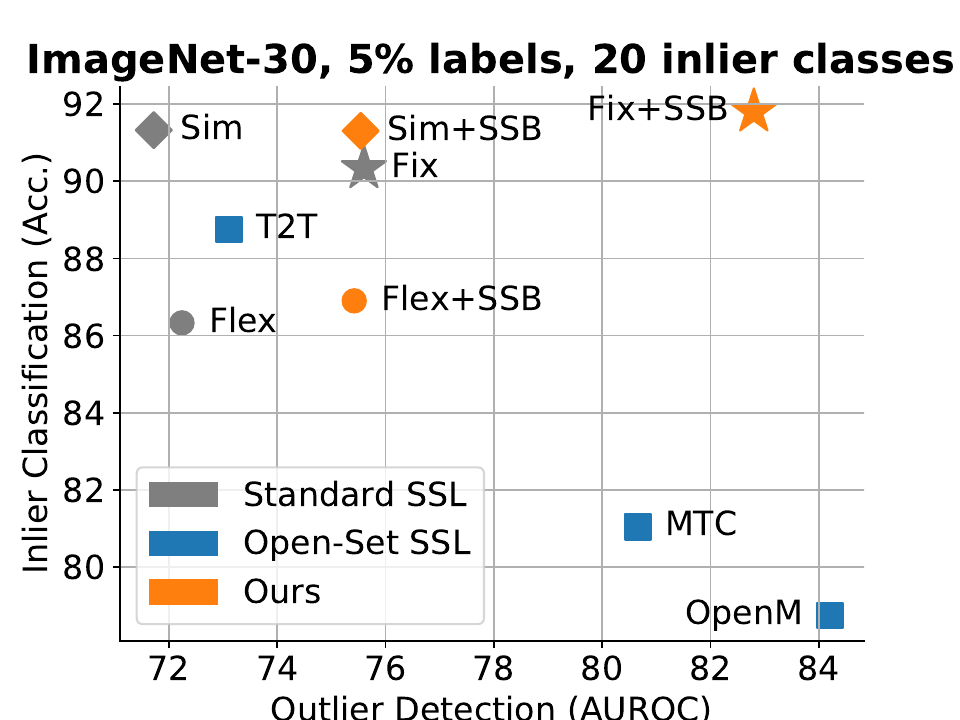}
\caption{
\textbf{Classification performance versus the outlier detection performance on ImageNet-30. }
\ourmethod achieves good performance in both inlier classification and OOD detection.
Please see Appendix A for a detailed breakdown of the results in tables.
}
\vspace{-7pt}
\label{fig:main_results_imagenet}
\end{figure}

\subsection{Main Results} \label{sec:main_results}
\myparagraph{Datasets \& Evaluation.}
As mentioned in Section \ref{sec:method}, the goal of open-set SSL is to train a good inlier classifier as well as an outlier detector that can identify both seen and unseen outliers.
Therefore, we need to construct three class spaces: inlier classes $\mathcal{C}$, seen outlier classes $\mathcal{U_S}$, and unseen outlier classes $\mathcal{U_U}$.
For each setting: the labeled set contains samples from $\mathcal{C}$ only; the unlabeled set contains samples from $\mathcal{C}$ and $\mathcal{U_S}$; the test set contains samples from $\mathcal{C}$, $\mathcal{U_S}$, and $\mathcal{U_U}$.
The inlier classification performance is evaluated on $\mathcal{C}$ using test accuracy as in standard supervised learning.
The OOD detection performance is measured by AUROC following \cite{saito2021openmatch} and we report the \textbf{average performance in detecting seen outliers and unseen outliers} (see Appendix A for separate AUROC on seen outliers and unseen outliers).

Following \cite{saito2021openmatch}, we evaluate \ourmethod on CIFAR-10 \cite{krizhevsky2009cifar}, CIFAR-100 \cite{krizhevsky2009cifar}, and ImageNet \cite{deng2009imagenet} with different numbers of labeled data.
For CIFAR-10, the 6 animal classes are used as inlier classes, and the rest 4 are used as seen outlier classes during the training.
Additionally, test sets from SVHN \cite{netzer2011svhn}, CIFAR-100, LSUN \cite{yu2015lsun}, and ImageNet are considered as unseen outliers, and used to evaluate the detection performance on unseen outliers.
For CIFAR-100, the inlier-outlier split is performed on super classes, and two settings are considered: 80 inlier classes (20 outlier classes) and 55 inlier classes (45 outlier classes).
Similar to CIFAR-10, test sets from SVHN, CIFAR-10, LSUN, and ImageNet are used to evaluate the detection performance on unseen outliers.
For ImageNet, we follow \cite{saito2021openmatch} to use ImageNet-30\cite{hendrycks2016baseline}, which is a subset of ImageNet containing 30 distinctive classes.
The first 20 classes are used as inlier classes while the rest 10 are used as outlier classes.
Stanford Dogs \cite{stanford_dogs}, CUB-200 \cite{cubuk2020randaugment}, Flowers102 \cite{nilsback2008flowers}, Caltech-256 \cite{griffin2007caltech256}, Describable Textures Dataset \cite{cimpoi14dtd}, LSUN are used as unseen outlier classes at test time.

\myparagraph{Implementation details.}
We use Wide ResNet-28-2 \cite{zagoruyko2016wrn} as the backbone for CIFAR experiments and ResNet-18 \cite{he2016resnet} for ImageNet experiments.
As standard SSL models do not have the notion of OOD detection, we adopt the method in \cite{hendrycks2016baseline}, where the OOD score of an input image $\textbf{x}$ is computed as 
$1 - \max \text{softmax}(f(\textbf{x}))$ and $f$ denotes the model.
Thus, the input image is considered as an outlier if the OOD score is higher than a pre-defined threshold.
For other open-set SSL methods, we directly employ the authors' implementations and follow their default hyper-parameters.

For SSB, we use two two-layer MLPs with ReLU \cite{nair2010relu} non-linearity to separate representations for all settings. 
The hidden dimension is 1024 for CIFAR settings and 4096 for ImageNet settings.
For classifier training, we follow \cite{sohn2020fixmatch} and set the threshold $\tau$ as 0.95.
For outlier detector training, we set $\lambda_{det}^u$ as 1 for all settings and follow \cite{saito2021openmatch} for the weights of OC loss and entropy minimization.
The threshold $\theta$ is 0.01 for all experiments \camrdy{(see ablation in Appendix C).}
Following \cite{saito2021openmatch}, we train our model for 512 epochs with SGD \cite{kiefer1952sgd} optimizer.
The learning rate is set as 0.03 with a cosine decay.
The batch size is 64.
Additionally, we defer the training of the outlier detector until epoch 475 to reduce the computational cost as we find empirically the deferred training does not comprise the model performance.
The ablation on the deferred training is in Appendix C.

When combined with standard SSL methods (e.g. SSB + FlexMatch), we replace the classifier training losses in Equation 1 with the corresponding losses of different methods while keeping the outlier detector the same.
When combined with open-set SSL methods (e.g. MTC + SSB), we make three modifications. First, we separate the outlier detector branch from the classifier branch using the proposed MLP projection head. Second, we replace the outlier detector training losses with our loss from Equation 6. Third, we do not filter unlabeled data with the outlier detector for classifier training.

\myparagraph{Results.}
We compare \ourmethod with both standard SSL and open-set SSL methods.
Fig. \ref{fig:main_results_cifar} and \ref{fig:main_results_imagenet} summarize the inlier test accuracy and outlier AUROC for CIFAR datasets and ImageNet, respectively.
\camrdy{Considering the goal of open-set SSL is to achieve \textit{both good inlier classification accuracy and outlier detection}, SSB greatly outperforms standard SSL methods in outlier detection, and open-set SSL methods in inlier classification.}
For example, on CIFAR-10 with 25 labels, the AUROC of our best method is 11.97\% higher than the best method excluding ours.
Moreover, when combined with standard SSL algorithms, our method demonstrates consistent improvement in OOD detection, and in most cases, better test accuracy for inlier classification. 
This suggests the flexibility of our method, which makes it possible to benefit from the most advanced approaches.
\camrdy{Note that the performance improvement of SSB can not be simply explained by the increased number of parameters introduced in the projection heads. Please see Fig. \ref{fig:fair_comparison} for a comparison between SSB and other methods + MLP heads.}

Additionally, \ourmethod is more robust to the number of labeled data than others.
We achieve reasonable performance given a small number of labeled data while other methods fail to generalize.
For example, on CIFAR-10 with 6 inlier classes, OpenMatch has similar inlier accuracy as ours at 50 labels.
When the number of labeled data is halved, their performance decreases to 54.88\% while our method still has a test accuracy of 91.74\%.
Please see Appendix B for comparisons on more benchmarks.

\collapser{
}

\subsection{Ablation Study} \label{sec:ablation}
In this section, we analyze the design choices of \ourmethod and show their importance through ablation experiments.
If not specified, we use CIFAR-10 with 25 labeled data as our default setting for ablation.
The same data split is used for fair comparison.

\myparagraph{Importance of non-linear projection heads.}
As mentioned in Section \ref{sec:method}, we use 2-layer MLPs to mitigate the adverse effect between the inlier classifier and outlier detector.
Here we study the effect of the projection heads in Table \ref{tab:disentangle}.
As we can see, incorporating confidence filtering yields a significant improvement in inlier classification performance (resulting in a 12.23\% to 13.18\% increase).
However, the OOD detection performance experiences a substantial decline when the projection heads are missing (AUROC from 89.67\% to 63.46\%).
This is because the classifier tends to mix the features of inliers and outliers with the same pseudo-labels in a shared feature space, which contradicts the goal of the outlier detector.
The addition of the projection heads not only restores the OOD detection performance but also achieves superior results when combined with confidence filtering.
Adding the projection heads in combination with confidence filtering not only restores the OOD detection performance but achieves even better performance, which indicates the importance of representation separation.
\camrdy{Note that it is important to have two independent projection heads for the inlier classifier and outlier detector. A shared projection head does not restore the OOD detection performance as shown in Table \ref{tab:disentangle}.}
Moreover, we show in Table \ref{tab:switch_head} that both classification and detection performance degrade when swapping the task-specific features of a pre-trained model with the fixed encoder. 
In particular, when re-training the detector (just a fully-connected layer) on top of classification features, seen AUROC drops from 89.18\% to 53.99\%, which suggests our model learns more task-specific features. 
Therefore, the utilization of the projection heads separates concerns between the classifier and detector, which eases the difficulties of the task and allows them to be trained jointly without adversely affecting each other. 
The effect of the depth and the width of the projection head is studied in Appendix C.



\begin{table}[ht]
\begin{center}
\begin{tabular}{cccc}
\toprule
\begin{tabular}[c]{@{}c@{}}Proj.\\ head\end{tabular} & \begin{tabular}[c]{@{}c@{}}Conf.\\ filter\end{tabular} & \begin{tabular}[c]{@{}c@{}}Inlier Cls.\\ (Acc.)\end{tabular} & \begin{tabular}[c]{@{}c@{}}Outlier Det.\\ (AUROC)\end{tabular} \\ \midrule
 &  & 78.05 & 89.67 \\
\camrdy{shared} &  & \camrdy{76.75} & \camrdy{91.92} \\
separate &  & 78.47 & 90.92 \\ \midrule
 & \checkmark & 90.28 & 63.46 \\
\camrdy{shared} & \camrdy{\checkmark} & \camrdy{90.93} & \camrdy{63.87} \\
separate & \checkmark & \textbf{91.65} & \textbf{94.76} \\ \bottomrule
\end{tabular}
\end{center}
\vspace{-5pt}
\caption{
\textbf{Effect of the projection head and confidence-based pseudo-labeling for classifier training.} We use a 2-layer MLP as the projection head. 
All models are trained with pseudo-negative mining on the same data split.
}
\label{tab:disentangle}
\end{table}

\begin{table}[ht]
\begin{center}
\vspace{5pt}
\includegraphics[width=0.8\linewidth]{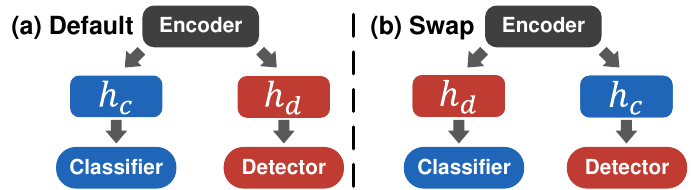}\\
\none{an empty line here}
\resizebox{0.9\linewidth}{!}{%
\begin{tabular}{ccc}
\toprule
Nearest Neighbor & \begin{tabular}[c]{@{}c@{}}Inlier Cls.\\ (Acc.)\end{tabular} & \begin{tabular}[c]{@{}c@{}}Outlier Det.\\ (AUROC)\end{tabular} \\ \midrule
default (a) & \textbf{55.04} & \textbf{99.43} \\
swap cls. \& det. features (b) & 53.70 & 77.89 \\ \bottomrule
\end{tabular}
}%
\end{center}
\vspace{-5pt}
\caption{
\textbf{Classification and detection performance using features of different heads.}
We fix the encoder and MLP heads and evaluate the classification and detection performance using nearest neighbors on labeled set.
Our model learns specialized features since swapping $h_c$ and $h_d$ leads to inferior performance in both tasks.
}
\vspace{-5pt}
\label{tab:switch_head}
\end{table}

\begin{figure}[t]
\centering
\includegraphics[width=0.95\linewidth]{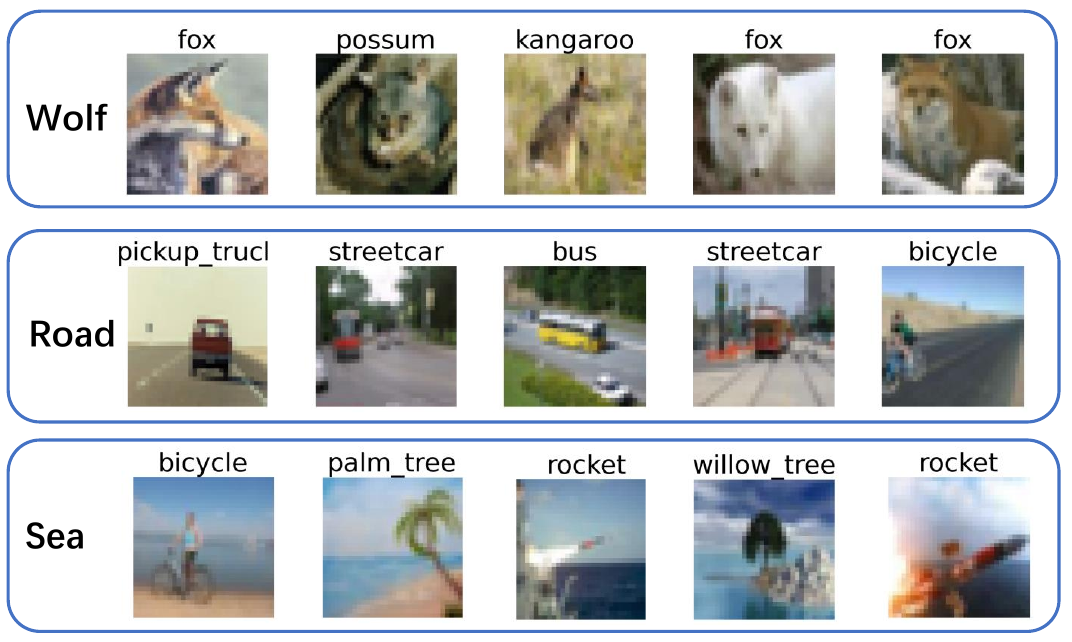}
\caption{
\textbf{OOD samples can be used as data augmentation to improve the generalization performance.}
The figure shows three semantic classes from labeled data (wolf, road, and sea), and top-5 confident OOD samples predicted for those classes.
The ground-truth semantic class of the OOD sample is on the top of each image.
We can see that OOD data with high confidence present large visual similarities to the corresponding semantic classes.
}
\vspace{-5pt}
\label{fig:ablation_ood}
\end{figure}

\myparagraph{Improving data utilization with confidence-based pseudo-labeling.}
Here we study the effect of different classifier training strategies. 
We compare three unlabeled data filtering methods for classifier training: (1) \textit{det.} selects pseudo-inliers with the outlier detector as in \cite{saito2021openmatch}; (2) \textit{det. (tuned)}, where we choose the selection threshold in detector-based filtering so that the recall of actual inlier samples matches ours; (3) \textit{conf.} uses unlabeled data whose confidence is higher than a pre-defined threshold, which is our method.
As shown in Table \ref{tab:clf_train}, although \textit{det.} successfully removes many OOD data, it also eliminates many inliers, resulting in a low utilization ratio of unlabeled data (0.29\% unlabeled data are used in training).
In contrast, our method includes pseudo-labels with high classifier confidence into the training, irrespective of whether a sample is out-of-distribution, which leads to a high utilization ratio of unlabeled data (94.22\%), thus, outperforming \textit{det.} with a large margin.
Moreover, our method also outperforms \textit{det (tuned)} whose data selection threshold is tuned for better performance.
This is because we incorporate a significant amount of OOD data in the training process (40.16\% v.s. 16.90\%). 
In fact, many OOD data are natural data augmentation of inliers, which can substantially improve closed-set classification if used carefully.
\camrdy{When removing pseudo-labeled OOD data using an oracle during the training. The inlier classification accuracy decreases by 3.37\% on CIFAR-10 with 25 labels (from 91.65\% to 88.28\%), which suggests pseudo-labeled OOD data are helpful for inlier classification.}
In Fig. \ref{fig:ablation_ood}, we visualize top-5 confident OOD samples predicted for three inlier classes from \textit{conf.} on CIFAR-100.
We can see that the selected samples are related to the inlier classes and contain the corresponding semantics despite being outliers.
For example, OOD data selected for \textit{sea} are images with sea background
(more examples in Appendix E).

\begin{table}[ht]
\begin{center}
\resizebox{\linewidth}{!}{%
\begin{tabular}{lllccc}
\toprule
\multicolumn{3}{l}{Filter method} & det. & det (tuned) & conf. (ours) \\ \midrule
\multirow{2}{*}{\rotatebox[origin=c]{90}{\textcolor{gray}{Test}}} & \multicolumn{2}{l}{Inlier Clf. (Acc.)} & 47.20 & 86.53 & \textbf{91.65} \\
 & \multicolumn{2}{l}{Outlier Det. (AUROC)} & 57.72 & 87.87 & \textbf{94.76} \\ \midrule
\multirow{5}{*}{\rotatebox[origin=c]{90}{\textcolor{gray}{Train}}} & \multicolumn{5}{l}{Utilization ratio of:} \\ 
 & \multicolumn{2}{l}{- Unlabeled} & 0.29 & 58.09 & 94.22 \\
 & \multicolumn{2}{l}{- OOD data} & 0.04 & 16.90 & 40.16 \\ \cmidrule{2-6}
 & \multicolumn{2}{l}{Prec. of pseudo-inliers} & 95.17 & 86.53 & 58.30 \\
 & \multicolumn{2}{l}{Recall of inliers} & 0.47 & 93.86 & 92.14 \\ \bottomrule
\end{tabular}
}%
\end{center}
\vspace{-5pt}
\caption{
\textbf{Effect of different OOD filtering methods for classifier training.}
We compare three filtering methods: \textit{conf.} denotes the confidence-based pseudo-labeling; \textit{det.} uses the outlier detector to select pseudo-inliers for classifier training; \textit{det. (tuned)} is a tuned version of \textit{det.} that matches the recall of inliers with our method.
We compare the performance as well as the data utilization ratio, precision, and recall of the inliers from unlabeled data during training.
All models are trained with pseudo-negative mining and the projection head on the same data split.
}
\label{tab:clf_train}
\end{table}


\begin{figure}[ht]
	\centering
	\begin{minipage}{0.5\linewidth}
	    \centering
	    \includegraphics[width=\linewidth]{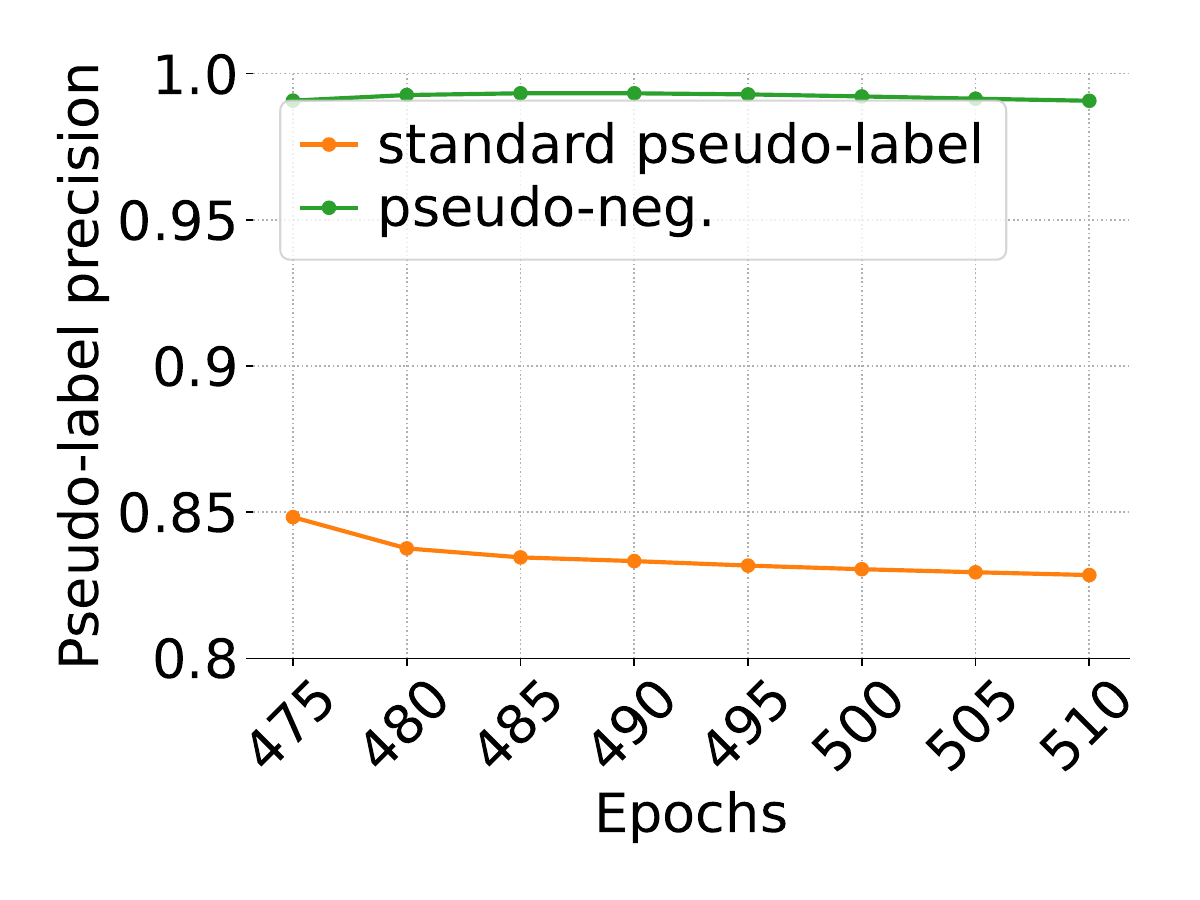}
	\end{minipage}
	\begin{minipage}{0.5\linewidth}
	    \centering
		\includegraphics[width=\linewidth]{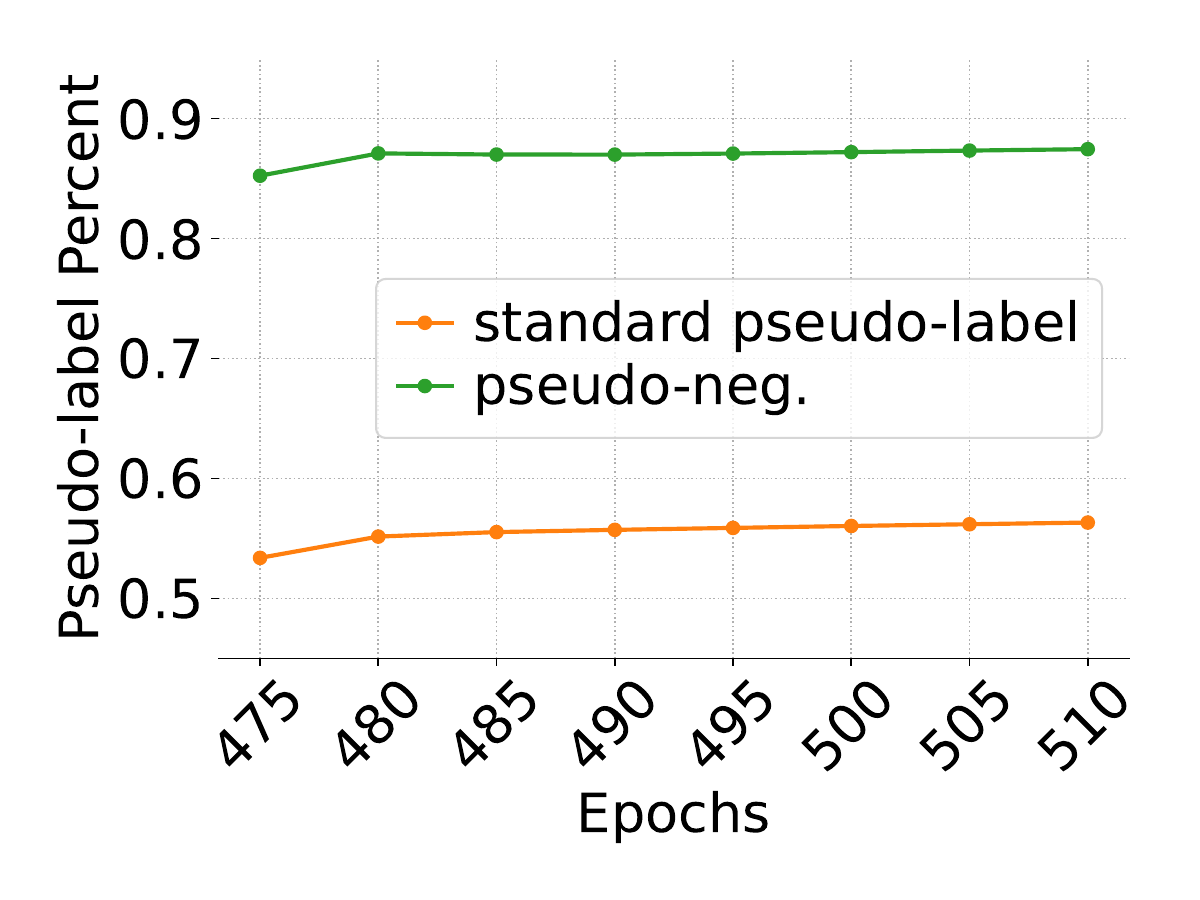}
	\end{minipage} 
\vspace{-5pt}
\caption{Compared to standard pseudo-labeling, pseudo-negative mining has not only higher prediction precision, but also higher data utilization rate.
}
\vspace{-5pt}
\label{fig:pseudo_neg_curve}
\end{figure}

\myparagraph{Effect of pseudo-negative mining.}
Table \ref{tab:ablation_pseudo_neg} shows the effect of pseudo-negative mining.
We compare our pseudo-negative mining with standard pseudo-labeling which predicts artificial labels for unlabeled data and uses confident predictions with labeled data loss.
While standard pseudo-labeling does not help the OOD detection performance further, pseudo-negative mining improves the seen AUROC by 4.73\% over the model without pseudo-negative mining.
Compared to standard pseudo-labeling, pseudo-negative mining not only includes more unlabeled data into the training, but also presents high precision for the selected pseudo-outliers as shown in Fig. \ref{fig:pseudo_neg_curve}.

As mentioned in Section \ref{sec:method_det}, we utilize unlabeled data with low inlier scores as pseudo-outliers to enhance the data diversity of outlier classes.
An unlabeled sample is used as a pseudo-outlier only if its confidence score is less than a pre-defined threshold $\theta$.
Table \ref{tab:pseudo_neg_theta} compares the results of different thresholds.
We can see that our method achieves similar performance as long as $\theta$ takes a relatively small value, which suggests the good robustness of our method against this hyper-parameter.
We provide more ablation on loss weight and data augmentation in Appendix C.

\begin{table}[ht]
\begin{center}
\resizebox{0.8\linewidth}{!}{%
\begin{tabular}{ccc}
\toprule
Threshold $\theta$ & \begin{tabular}[c]{@{}c@{}}Inlier Cls.\\ (Acc.)\end{tabular}  & \begin{tabular}[c]{@{}c@{}}Outlier Det.\\ (seen AUROC)\end{tabular} \\ \midrule
0.2 & 91.87 & 92.96 \\
0.1 & \textbf{92.03} & 93.16 \\
0.05 & 91.97 & 94.21 \\
0.01 & 91.65 & \textbf{94.76} \\
0.005 & 91.52 & 94.75 \\
0.001 & 91.70 & 94.15 \\
\bottomrule
\end{tabular}
}%
\end{center}
\vspace{-5pt}
\caption{
\textbf{Effect of different thresholds $\theta$ for pseudo-negative mining.}
Our method shows good robustness against a wide range of thresholds.
We use CIFAR-10 with 25 labeled data here.
}
\label{tab:pseudo_neg_theta}
\end{table}

\begin{table}[ht]
\begin{center}
\begin{tabular}{lcc}
\toprule
\begin{tabular}[c]{@{}l@{}}Pseudo-\\ labeling\end{tabular} & \begin{tabular}[c]{@{}c@{}}Inlier Cls.\\ (Acc.)\end{tabular} & \begin{tabular}[c]{@{}c@{}}Outlier Det.\\ (AUROC)\end{tabular} \\ \midrule
None & 91.52 & 90.03 \\
Standard & 91.63 & 89.69 \\
Pseudo-neg. & \textbf{91.65} & \textbf{94.76} \\ \bottomrule
\end{tabular}
\end{center}
\vspace{-5pt}
\caption{
\textbf{Effect of pseudo-negative mining for OOD detection.}
All models are trained with confidence-based pseudo-labeling and a 2-layer MLP projection head on the same data split.
}
\vspace{-5pt}
\label{tab:ablation_pseudo_neg}
\end{table}

\myparagraph{Ablation on outlier detectors.}
Here, we compare the performance of different outlier detection methods. 
Specifically, we choose three schemes from recent works, including the binary classifier from MTC \cite{yu2020mtc}, cross-modal matching from T2T \cite{huang2021t2t}, and OVA classifiers from OpenMatch \cite{saito2021openmatch}. 
As shown in Table \ref{tab:ablation_det}.
While all methods show  reasonable performance, OVA classifiers exhibit the best performance in both inlier classification and OOD detection.
Hence, we use OVA classifiers as the outlier detector in our final model.

\begin{table}[ht]
\begin{center}
\resizebox{\linewidth}{!}{%
\begin{tabular}{lcc}
\toprule
OOD Detector & \begin{tabular}[c]{@{}c@{}}Inlier Cls.\\ (Acc.)\end{tabular} & \begin{tabular}[c]{@{}c@{}}Outlier Det.\\ (AUROC)\end{tabular} \\ \midrule
binary classifier~\cite{yu2020mtc} & 70.93 & 76.12 \\
cross-modal matching~\cite{huang2021t2t} & 69.27 & 75.99 \\
OVA classifiers~\cite{saito2021openmatch} & \textbf{71.00} & \textbf{82.62} \\ \bottomrule
\end{tabular}
}%
\end{center}
\vspace{-5pt}
\caption{
\textbf{Comparison between different outlier detectors.}
The experiment is conducted on CIFAR-100 with 55 inlier classes and 25 labels per class.
}
\vspace{-5pt}
\label{tab:ablation_det}
\end{table}

\myparagraph{Compatibility with other open-set SSL methods.}
We evaluated the compatibility of our method with other open-set SSL techniques in Table \ref{tab:ablation_flex}. 
Our results indicate that our method is highly compatible, as all existing methods showed improved performance in both inlier classification and outlier detection when combined with our approach. 
This demonstrates the flexibility of our method and suggests that it can be easily integrated into existing frameworks as a plug-and-play solution.



\begin{table}[ht]
\begin{center}
\begin{tabular}{lcc}
\toprule
 & \begin{tabular}[c]{@{}c@{}}Inlier Cls.\\ (Acc.)\end{tabular} & \begin{tabular}[c]{@{}c@{}}Outlier Det.\\ (AUROC)\end{tabular} \\ \midrule
MTC & 60.24 & 69.88 \\
MTC + Ours & \textbf{60.42} & \textbf{74.98} \\ \midrule
T2T & 64.78 & 52.93 \\
T2T + Ours & \textbf{66.98} & \textbf{69.50} \\ \midrule
OpenMatch & 68.53 & 80.00 \\
OpenM. + Ours & \textbf{71.00} & \textbf{82.62} \\ \bottomrule
\end{tabular}
\end{center}
\vspace{-5pt}
\caption{
\textbf{Integrating our method with other open-set SSL methods improves performance.}
The setting is CIFAR-100 with 55 inlier classes and 25 labels per class.
}
\label{tab:ablation_flex}
\end{table}

\begin{figure}[ht]
\centering
\begin{minipage}{0.8\linewidth}
    \centering
    \includegraphics[width=\textwidth]{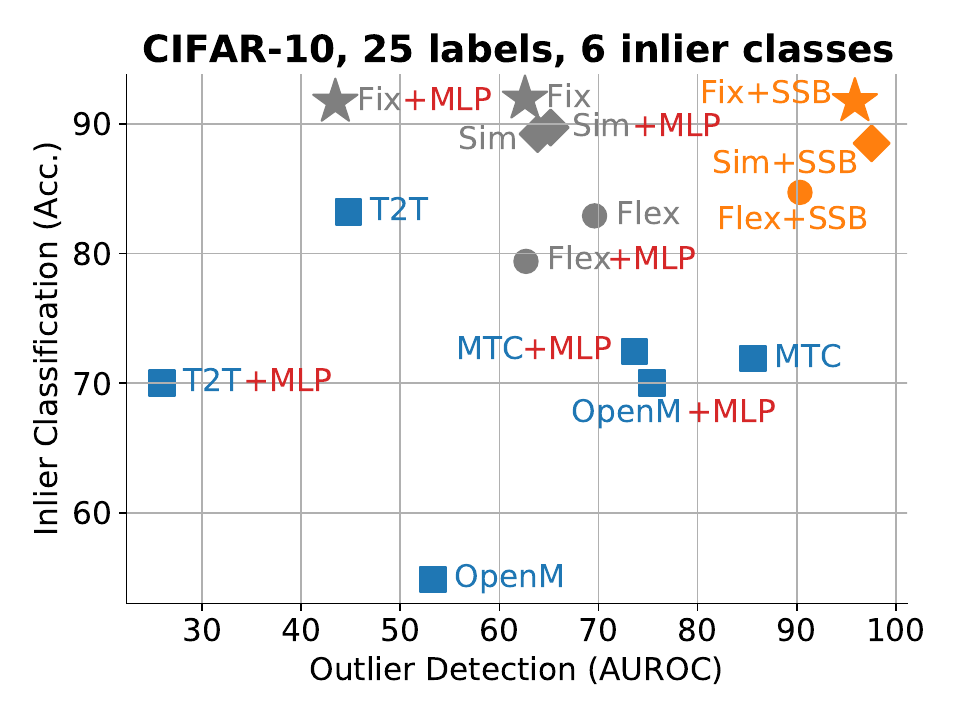}
\end{minipage}
\begin{minipage}{0.8\linewidth}
    \centering
    \includegraphics[width=\textwidth]{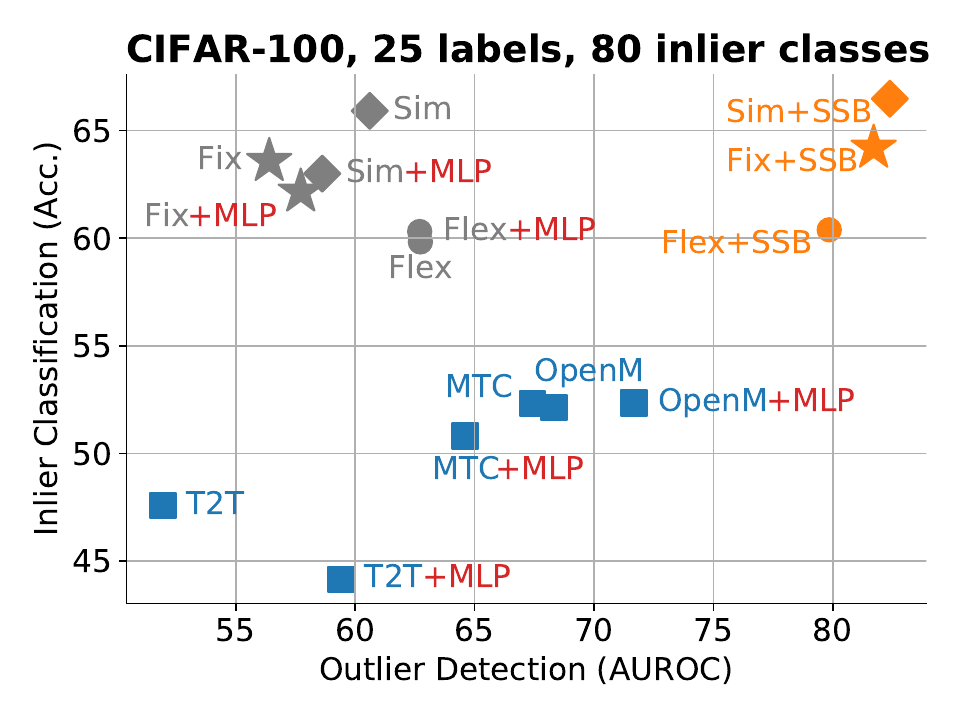}
\end{minipage}%
\vspace{-5pt}
\caption{
\textbf{Comparison between SSB and other methods with the same model parameters.}
The performance improvement of SSB can not be simply explained by the increased number of parameters.
}
\vspace{-5pt}
\label{fig:fair_comparison}
\end{figure}

\myparagraph{Equal-parameter comparison.}
As mentioned in Section \ref{sec:main_results}, the performance improvement of SSB can not be simply explained by the increased number of parameters introduced in the projection heads.
Here we compare SSB with other methods + MLP heads so that they have the same number of parameters as SSB.
As shown in Fig. \ref{fig:fair_comparison}, adding MLP heads improves the  performance of other methods, but SSB still greatly outperforms all of them, indicating that the performance improvement of our method can not be merely explained by the increase of the model capacity.

\section{Conclusion and Limitations}
In this paper, we study a realistic and challenging setting, open-set SSL, where unlabeled data contains outliers from categories that do not appear in the labeled data.
We first demonstrate that classifier-confidence-based pseudo-labeling can effectively improve the unlabeled data utilization ratio and leverage useful OOD data, which largely improves the classification performance.
We find that adding non-linear transformations between the task-specific head and the shared features provides sufficient decoupling of the two heads, which prevents mutual interference and improves performance in both tasks.
Additionally, we propose pseudo-negative mining to improve OOD detection. It uses pseudo-outliers to enhance the representation learning of OOD data, which further improves the model's ability to distinguish between inliers and OOD samples.
Overall, we achieve state-of-the-art performance on several benchmark datasets, demonstrating the effectiveness of the proposed method.

Nonetheless, \ourmethod has potential limitations.
Despite the improved overall performance, the outlier detector suffers from overfitting as the performance gap between detecting seen outliers and unseen outliers is still very large. 
Therefore, in the future, more regularizations need to be considered to improve the generalization.
Another drawback is that our method is not able to deal with long-tail distributions, which is also very realistic in practice. 
Presumably, our method will have difficulty distinguishing inliers of tail classes and OOD data due to the data scarcity at tail.

{\small
\bibliographystyle{ieee_fullname}
\bibliography{egbib}
}
\newpage
\appendix

In this appendix, we first show the tabulated breakdown of our main result in Section \ref{sec:breakdown}.
Then we compare our method with more recent approaches in Section \ref{sec:benchmark} and show additional ablation studies in Section \ref{sec:ablation_appendix}.
Finally, we present the pseudo-code and more visualizations of pseudo-inliers in Section \ref{sec:pseudo_code} and Section \ref{sec:visualization}, respectively.

\section{Main Results Breakdown} \label{sec:breakdown}
Here we present tabulated breakdown results of Fig. 3 and 4 from the main paper.
We summarize the inlier classification accuracy and outlier detection in AUROC for different settings in Table \ref{tab:main_1}, \ref{tab:main_2}, \ref{tab:main_3}, \ref{tab:main_4}, \ref{tab:main_5}, \ref{tab:main_6}, and \ref{tab:main_7}, respectively.
To provide a more comprehensive analysis, we further provide the separate outlier detection performance results for seen outliers and unseen outliers in Table \ref{tab:seen_auroc} and Table \ref{tab:unseen_auroc}, respectively.
SSB achieves competitive results in all settings.
In particular, for CIFAR-10 and CIFAR-100 with 25 labels, SSB outperforms other methods by a large margin.


\begin{table}[ht]
\begin{center}
\begin{tabular}{llc}
\toprule
\multicolumn{2}{l}{Test Acc. / AUROC} & CIFAR-10 \\ \cmidrule(l){1-2} \cmidrule(l){3-3}
\multicolumn{2}{l}{inlier / outlier classes} & 6 / 4 \\ \cmidrule(l){1-2} \cmidrule(l){3-3}
\multicolumn{2}{l}{labels per class} & 25 \\ \midrule
\multirow{3}{*}{\rotatebox[origin=c]{90}{\textcolor{gray}{SSL}}} & FixMatch~\cite{sohn2020fixmatch} & $\textbf{91.94}_{\pm0.16}$ / $62.58_{\pm0.53}$  \\ 
 & FlexMatch~\cite{zhang2021flexmatch} & $82.91_{\pm0.92}$ / $69.60_{\pm4.11}$ \\
 & SimMatch~\cite{zheng2022simmatch} & $89.22_{\pm2.24}$ / $63.85_{\pm0.70}$ \\ \midrule
\multirow{3}{*}{\rotatebox[origin=c]{90}{\textcolor{gray}{OSSL}}} & MTC~\cite{yu2020mtc} & $71.91_{\pm10.82}$ / $85.57_{\pm6.63}$ \\
 & OpenMatch~\cite{saito2021openmatch} & $54.88_{\pm2.33}$ / $53.32_{\pm4.62}$ \\
 & T2T~\cite{huang2021t2t} & $83.21_{\pm0.98}$ / $44.79_{\pm17.26}$ \\ \midrule
 & SSB (FixMatch) & \underline{91.74}$_{\pm0.24}$ / $\underline{95.86}_{\pm1.37}$ \\
 & SSB (FlexMatch) & $84.72_{\pm0.49}$ / $90.32_{\pm0.75}$ \\
 & SSB (SimMatch) & $88.51_{\pm2.86}$ / $\textbf{97.54}_{\pm0.08}$ \\
 \bottomrule
\end{tabular}
\end{center}
\vspace{-5pt}
\caption{
\textbf{CIFAR-10 with 25 labels and 6 inlier classes.}
We report test accuracy (\%) / AUROC (\%) for inliers classification and outlier detection, respectively.
The numbers are averaged over 3 different random seeds. 
The best number is in \textbf{bold}, and the second best is in \underline{underline}.
}
\vspace{-5pt}
\label{tab:main_1}
\end{table}

\newpage

\begin{table}[ht]
\begin{center}
\begin{tabular}{llc}
\toprule
\multicolumn{2}{l}{Test Acc. / AUROC} & CIFAR-10 \\ \cmidrule(l){1-2} \cmidrule(l){3-3}
\multicolumn{2}{l}{inlier / outlier classes} & 6 / 4 \\ \cmidrule(l){1-2} \cmidrule(l){3-3}
\multicolumn{2}{l}{labels per class} & 50 \\ \midrule
\multirow{3}{*}{\rotatebox[origin=c]{90}{\textcolor{gray}{SSL}}} & FixMatch~\cite{sohn2020fixmatch} & $\underline{91.33}_{\pm0.18}$ / $63.77_{\pm0.14}$ \\ 
 & FlexMatch~\cite{zhang2021flexmatch} & $83.98_{\pm0.31}$ / $64.47_{\pm0.10}$ \\ 
 & SimMatch~\cite{zheng2022simmatch} & $91.10_{\pm0.52}$ / $65.34_{\pm0.09}$ \\ \midrule
\multirow{3}{*}{\rotatebox[origin=c]{90}{\textcolor{gray}{OSSL}}} & MTC~\cite{yu2020mtc} & $81.03_{\pm5.21}$ / $92.01_{\pm2.62}$ \\
 & OpenMatch~\cite{saito2021openmatch} & $91.31_{\pm1.18}$ / $\underline{95.88}_{\pm0.60}$ \\
 & T2T~\cite{huang2021t2t} & $90.56_{\pm0.07}$ / $39.73_{\pm8.94}$ \\ \midrule
 & SSB (FixMatch) & $\textbf{92.18}_{\pm0.33}$ / $\textbf{97.65}_{\pm0.19}$ \\ 
 & SSB (FlexMatch) & $84.26_{\pm1.36}$ / $93.16_{\pm3.63}$ \\ 
 & SSB (SimMatch) & $90.82_{\pm0.47}$ / $94.07_{\pm0.40}$ \\ 
 \bottomrule
\end{tabular}
\end{center}
\vspace{-5pt}
\caption{
\textbf{CIFAR-10 with 50 labels and 6 inlier classes.}
We report test accuracy (\%) / AUROC (\%) for inliers classification and outlier detection, respectively.
The numbers are averaged over 3 different random seeds. 
The best number is in \textbf{bold}, and the second best is in \underline{underline}.
}
\vspace{-5pt}
\label{tab:main_2}
\end{table}

\begin{table}[H]
\begin{center}
\vspace{-10pt}
\begin{tabular}{llc}
\toprule
\multicolumn{2}{l}{Test Acc. / AUROC} & CIFAR-100 \\ \cmidrule(l){1-2} \cmidrule(l){3-3}
\multicolumn{2}{l}{inlier / outlier classes} & 55 / 45 \\ \cmidrule(l){1-2} \cmidrule(l){3-3}
\multicolumn{2}{l}{labels per class} & 25 \\ \midrule
\multirow{3}{*}{\rotatebox[origin=c]{90}{\textcolor{gray}{SSL}}} & FixMatch~\cite{sohn2020fixmatch} & $69.89_{\pm0.00}$ / $63.81_{\pm0.26}$ \\ 
 & FlexMatch~\cite{zhang2021flexmatch} & $67.87_{\pm0.58}$ / $67.65_{\pm1.12}$ \\ 
 & SimMatch~\cite{zheng2022simmatch} & $70.25_{\pm0.96}$ / $65.21_{\pm0.68}$ \\ \midrule
\multirow{3}{*}{\rotatebox[origin=c]{90}{\textcolor{gray}{OSSL}}} & MTC~\cite{yu2020mtc} & $58.13_{\pm2.11}$ / $71.62_{\pm1.36}$ \\
 & OpenMatch~\cite{saito2021openmatch} & $67.09_{\pm1.44}$ / $80.18_{\pm0.09}$ \\
 & T2T~\cite{huang2021t2t} & $65.71_{\pm0.93}$ / $60.11_{\pm6.25}$ \\ \midrule
 & SSB (FixMatch) & $\underline{70.64}_{\pm0.36}$ / $82.91_{\pm0.30}$ \\ 
 & SSB (FlexMatch) & $68.28_{\pm0.74}$ / $\underline{83.62}_{\pm0.43}$ \\ 
 & SSB (SimMatch) & $\textbf{70.77}_{\pm0.54}$ / $\textbf{84.77}_{\pm0.52}$ \\ 
 \bottomrule
\end{tabular}
\end{center}
\vspace{-5pt}
\caption{
\textbf{CIFAR-100 with 25 labels and 55 inlier classes.}
We report test accuracy (\%) / AUROC (\%) for inliers classification and outlier detection, respectively.
The numbers are averaged over 3 different random seeds. 
The best number is in \textbf{bold}, and the second best is in \underline{underline}.
}
\vspace{-8pt}
\label{tab:main_3}
\end{table}

\newpage

\begin{table}[ht]
\begin{center}
\begin{tabular}{llc}
\toprule
\multicolumn{2}{l}{Test Acc. / AUROC} & CIFAR-100 \\ \cmidrule(l){1-2} \cmidrule(l){3-3}
\multicolumn{2}{l}{inlier / outlier classes} & 55 / 45 \\ \cmidrule(l){1-2} \cmidrule(l){3-3}
\multicolumn{2}{l}{labels per class} & 50 \\ \midrule
\multirow{3}{*}{\rotatebox[origin=c]{90}{\textcolor{gray}{SSL}}} & FixMatch~\cite{sohn2020fixmatch} & $73.28_{\pm0.59}$ / $66.03_{\pm0.41}$ \\ 
 & FlexMatch~\cite{zhang2021flexmatch} & $71.74_{\pm0.01}$ / $70.36_{\pm0.60}$ \\ 
 & SimMatch~\cite{zheng2022simmatch} & $\underline{74.15}_{\pm0.57}$ / $67.34_{\pm0.19}$ \\ \midrule
\multirow{3}{*}{\rotatebox[origin=c]{90}{\textcolor{gray}{OSSL}}} & MTC~\cite{yu2020mtc} & $65.97_{\pm0.77}$ / $68.96_{\pm1.08}$ \\
 & OpenMatch~\cite{saito2021openmatch} & $71.90_{\pm1.05}$ / $82.82_{\pm0.47}$ \\
 & T2T~\cite{huang2021t2t} & $70.69_{\pm0.11}$ / $61.97_{\pm0.50}$ \\ \midrule
 & SSB (FixMatch) & $73.70_{\pm0.75}$ / $\underline{85.89}_{\pm0.07}$ \\ 
 & SSB (FlexMatch) & $72.65_{\pm0.25}$ / $\textbf{85.97}_{\pm0.46}$ \\ 
 & SSB (SimMatch) & $\textbf{75.15}_{\pm0.34}$ / $84.60_{\pm0.18}$ \\ 
 \bottomrule
\end{tabular}
\end{center}
\caption{
\textbf{CIFAR-100 with 50 labels and 55 inlier classes.}
We report test accuracy (\%) / AUROC (\%) for inliers classification and outlier detection, respectively.
The numbers are averaged over 3 different random seeds. 
The best number is in \textbf{bold}, and the second best is in \underline{underline}.
}
\label{tab:main_4}
\end{table}

\begin{table}[ht]
\begin{center}
\begin{tabular}{llc}
\toprule
\multicolumn{2}{l}{Test Acc. / AUROC} & CIFAR-100 \\ \cmidrule(l){1-2} \cmidrule(l){3-3}
\multicolumn{2}{l}{inlier / outlier classes} & 80 / 20 \\ \cmidrule(l){1-2} \cmidrule(l){3-3}
\multicolumn{2}{l}{labels per class} & 25 \\ \midrule
\multirow{3}{*}{\rotatebox[origin=c]{90}{\textcolor{gray}{SSL}}} & FixMatch~\cite{sohn2020fixmatch} & $63.58_{\pm0.36}$ / $56.40_{\pm0.21}$ \\ 
 & FlexMatch~\cite{zhang2021flexmatch} & $59.83_{\pm1.78}$ / $62.73_{\pm0.62}$ \\ 
 & SimMatch~\cite{zheng2022simmatch} & $\underline{65.92}_{\pm0.81}$ / $60.61_{\pm0.60}$ \\ \midrule
\multirow{3}{*}{\rotatebox[origin=c]{90}{\textcolor{gray}{OSSL}}} & MTC~\cite{yu2020mtc} & $52.32_{\pm0.13}$ / $67.43_{\pm0.38}$ \\
 & OpenMatch~\cite{saito2021openmatch} & $52.13_{\pm4.81}$ / $68.32_{\pm4.68}$ \\
 & T2T~\cite{huang2021t2t} & $47.58_{\pm10.38}$ / $51.95_{\pm4.44}$ \\ \midrule
 & SSB (FixMatch) & $64.20_{\pm0.41}$ / $\underline{81.71}_{\pm0.86}$ \\ 
 & SSB (FlexMatch) & $60.39_{\pm1.89}$ / $79.85_{\pm0.94}$ \\ 
 & SSB (SimMatch) & $\textbf{66.48}_{\pm0.77}$ / $\textbf{82.39}_{\pm2.97}$ \\ 
 \bottomrule
\end{tabular}
\end{center}
\caption{
\textbf{CIFAR-100 with 25 labels and 80 inlier classes.}
We report test accuracy (\%) / AUROC (\%) for inliers classification and outlier detection, respectively.
The numbers are averaged over 3 different random seeds. 
The best number is in \textbf{bold}, and the second best is in \underline{underline}.
}
\label{tab:main_5}
\end{table}

\begin{table}[ht]
\begin{center}
\begin{tabular}{llc}
\toprule
\multicolumn{2}{l}{Test Acc. / AUROC} & CIFAR-100 \\ \cmidrule(l){1-2} \cmidrule(l){3-3}
\multicolumn{2}{l}{inlier / outlier classes} & 80 / 20 \\ \cmidrule(l){1-2} \cmidrule(l){3-3}
\multicolumn{2}{l}{labels per class} & 50 \\ \midrule
\multirow{3}{*}{\rotatebox[origin=c]{90}{\textcolor{gray}{SSL}}} & FixMatch~\cite{sohn2020fixmatch} & $67.06_{\pm0.10}$ / $58.05_{\pm0.49}$ \\ 
 & FlexMatch~\cite{zhang2021flexmatch} & $65.22_{\pm0.18}$ / $65.00_{\pm0.07}$ \\ 
 & SimMatch~\cite{zheng2022simmatch} & $\underline{69.35}_{\pm0.26}$ / $61.44_{\pm0.16}$ \\ \midrule
\multirow{3}{*}{\rotatebox[origin=c]{90}{\textcolor{gray}{OSSL}}} & MTC~\cite{yu2020mtc} & $59.17_{\pm0.01}$ / $69.34_{\pm1.81}$ \\
 & OpenMatch~\cite{saito2021openmatch} & $66.90_{\pm0.19}$ / $79.95_{\pm0.26}$ \\
 & T2T~\cite{huang2021t2t} & $64.18_{\pm0.64}$ / $65.26_{\pm13.73}$ \\ \midrule
 & SSB (FixMatch) & $67.97_{\pm0.20}$ / $80.81_{\pm1.02}$ \\ 
 & SSB (FlexMatch) & $65.79_{\pm0.06}$ / $\textbf{83.32}_{\pm0.36}$ \\ 
 & SSB (SimMatch) & $\textbf{70.27}_{\pm0.19}$ / $\underline{81.16}_{\pm2.10}$ \\ 
 \bottomrule
\end{tabular}
\end{center}
\caption{
\textbf{CIFAR-100 with 50 labels and 80 inlier classes.}
We report test accuracy (\%) / AUROC (\%) for inliers classification and outlier detection, respectively.
The numbers are averaged over 3 different random seeds. 
The best number is in \textbf{bold}, and the second best is in \underline{underline}.
}
\label{tab:main_6}
\end{table}

\begin{table}[ht]
\centering
\begin{center}
\begin{tabular}{llc}
\toprule
\multicolumn{2}{l}{Test Acc. / AUROC} & \multicolumn{1}{c}{ImageNet-30} \\ \cmidrule(l){1-2} \cmidrule(l){3-3} 
\multicolumn{2}{l}{inlier / outlier classes} & \multicolumn{1}{c}{20 / 10} \\ \cmidrule(l){1-2} \cmidrule(l){3-3}
\multicolumn{2}{l}{labels per class} & 5\% \\ \midrule
\multirow{3}{*}{\rotatebox[origin=c]{90}{\textcolor{gray}{SSL}}} & FixMatch~\cite{sohn2020fixmatch} & $90.33_{\pm0.66}$ / $75.60_{\pm1.28}$ \\ 
 & FlexMatch~\cite{zhang2021flexmatch} & $86.33_{\pm0.92}$ / $72.24_{\pm0.45}$ \\ 
 & SimMatch~\cite{zheng2022simmatch} & $\underline{91.32}_{\pm0.73}$ / $71.72_{\pm0.14}$ \\ \midrule
\multirow{3}{*}{\rotatebox[origin=c]{90}{\textcolor{gray}{OSSL}}} & MTC~\cite{yu2020mtc} & $81.05_{\pm0.35}$ / $80.66_{\pm2.23}$\\
 & OpenMatch~\cite{saito2021openmatch} & $78.75_{\pm0.35}$ / $\textbf{84.21}_{\pm0.03}$ \\
 & T2T~\cite{huang2021t2t} & $88.75_{\pm0.90}$ / $73.11_{\pm1.11}$\\ \midrule
 & SSB (FixMatch) & $\textbf{91.80}_{\pm0.05}$ / $\underline{82.80}_{\pm1.18}$ \\ 
 & SSB (FlexMatch) & $86.90_{\pm0.70}$ / $75.42_{\pm0.24}$ \\ 
 & SSB (SimMatch) & $91.30_{\pm0.65}$ / $75.54_{\pm0.10}$ \\ 
 \bottomrule
\end{tabular}
\end{center}
\caption{
\textbf{ImageNet-30 with 5\% labels and 20 inliers classes.}
We report test accuracy (\%) / AUROC (\%) for inliers classification and outlier detection, respectively.
The numbers are averaged over 3 different random seeds. 
The best number is in \textbf{bold}, and the second best is in \underline{underline}.
}
\label{tab:main_7}
\end{table}

\begin{table*}[ht]
\begin{center}
\begin{tabular}{llccccccc}
\toprule
\multicolumn{2}{l}{Seen AUORC} & \multicolumn{2}{c}{CIFAR-10} & \multicolumn{2}{c}{CIFAR-100} & \multicolumn{2}{c}{CIFAR-100} & ImageNet-30 \\ \cmidrule(l){1-2} \cmidrule(l){3-4} \cmidrule(l){5-6} \cmidrule(l){7-8} \cmidrule(l){9-9} 
\multicolumn{2}{l}{Inlier / outlier classes} & \multicolumn{2}{c}{6 / 4} & \multicolumn{2}{c}{55 / 45} & \multicolumn{2}{c}{80 / 20} & 20 / 10 \\ \cmidrule(l){1-2} \cmidrule(l){3-4} \cmidrule(l){5-6} \cmidrule(l){7-8} \cmidrule(l){9-9} 
\multicolumn{2}{l}{Labels per class} & 25 & 50 & 25 & 50 & 25 & 50 & 5\% \\ \midrule
\multirow{3}{*}{\rotatebox[origin=c]{90}{\textcolor{gray}{SSL}}} & FixMatch~\cite{sohn2020fixmatch} & $37.37_{\pm0.84}$ & $39.41_{\pm0.15}$ & $54.48_{\pm1.05}$ & $55.77_{\pm0.95}$ & $41.43_{\pm0.10}$ & $44.33_{\pm0.79}$ & $65.09_{\pm2.09}$ \\ 
& FlexMatch~\cite{zhang2021flexmatch} & $51.32_{\pm8.24}$ & $41.01_{\pm0.20}$ & $60.82_{\pm1.07}$ & $63.72_{\pm1.15}$ & $53.68_{\pm0.48}$ & $57.45_{\pm0.70}$ & $61.73_{\pm0.31}$ \\
& SimMatch~\cite{zheng2022simmatch} & $38.39_{\pm0.87}$ & $41.15_{\pm0.26}$ & $55.28_{\pm1.35}$ & $57.69_{\pm0.16}$ & $ 49.19_{\pm0.43}$ & $49.53_{\pm0.26}$ & $56.41_{\pm0.55}$ \\ \midrule
\multirow{3}{*}{\rotatebox[origin=c]{90}{\textcolor{gray}{OSSL}}} & MTC~\cite{yu2020mtc} & $92.00_{\pm3.49}$ & $94.47_{\pm2.05}$ & $76.93_{\pm1.48}$ & $72.53_{\pm0.18}$ & $69.15_{\pm0.86}$ & $72.38_{\pm1.86}$ & $82.70_{\pm2.60}$ \\
 & OpenMatch~\cite{saito2021openmatch} & $62.46_{\pm4.19}$ & $\underline{99.41}_{\pm0.18}$ & $84.93_{\pm0.08}$ & $86.99_{\pm0.23}$ & $74.87_{\pm3.78}$ & $\underline{86.19}_{\pm0.48}$ & $\textbf{91.79}_{\pm0.49}$ \\
 & T2T~\cite{huang2021t2t} & $34.90_{\pm27.50}$ & $23.85_{\pm8.45}$ & $52.95_{\pm6.15}$ & $59.50_{\pm1.50}$ & $50.45_{\pm9.15}$ & $61.40_{\pm21.10}$ & $62.35_{\pm2.05}$ \\ \midrule
 & SSB (FixMatch) & $\underline{99.35}_{\pm0.38}$ & $\textbf{99.63}_{\pm0.15}$ & $89.39_{\pm0.44}$ & $\underline{90.62}_{\pm0.46}$ & $\textbf{90.25}_{\pm1.34}$ & $85.29_{\pm2.18}$ & $\underline{83.84}_{\pm2.31}$ \\
 & SSB (FlexMatch) & $96.81_{\pm0.25}$ & $93.41_{\pm6.22}$ & $\textbf{89.84}_{\pm0.05}$ & $\textbf{91.21}_{\pm0.26}$ & $88.32_{\pm1.36}$ & $\textbf{90.68}_{\pm0.27}$ & $67.58_{\pm0.55}$ \\ 
 & SSB (SimMatch) & $\textbf{99.61}_{\pm0.10}$ & $93.07_{\pm0.70}$ & $\underline{89.75}_{\pm0.90}$ & $87.38_{\pm0.13}$ & $\underline{88.65}_{\pm3.86}$ & $83.05_{\pm3.49}$ & $62.63_{\pm0.09}$ \\ \bottomrule
\end{tabular}
\end{center}
\caption{
\textbf{AUROC (\%) for seen outliers.}
The best number is in \textbf{bold}, and the second best is in \underline{underline}.
Note that a random OOD detector gives an AUROC of 50\%.
}
\label{tab:seen_auroc}
\end{table*}

\begin{table*}[ht]
\begin{center}
\begin{tabular}{llccccccc}
\toprule
\multicolumn{2}{l}{Unseen AUROC} & \multicolumn{2}{c}{CIFAR-10} & \multicolumn{2}{c}{CIFAR-100} & \multicolumn{2}{c}{CIFAR-100} & ImageNet-30 \\ \cmidrule(l){1-2} \cmidrule(l){3-4} \cmidrule(l){5-6} \cmidrule(l){7-8} \cmidrule(l){9-9} 
\multicolumn{2}{l}{Inlier / outlier classes} & \multicolumn{2}{c}{6 / 4} & \multicolumn{2}{c}{55 / 45} & \multicolumn{2}{c}{80 / 20} & 20 / 10 \\ \cmidrule(l){1-2} \cmidrule(l){3-4} \cmidrule(l){5-6} \cmidrule(l){7-8} \cmidrule(l){9-9} 
\multicolumn{2}{l}{Labels per class} & 25 & 50 & 25 & 50 & 25 & 50 & 5\% \\ \midrule
\multirow{3}{*}{\rotatebox[origin=c]{90}{\textcolor{gray}{SSL}}} & FixMatch~\cite{sohn2020fixmatch} & $87.80_{\pm0.23}$ & $88.12_{\pm0.42}$ & $73.15_{\pm0.53}$ & $76.28_{\pm0.13}$ & $71.37_{\pm0.32}$ & $71.77_{\pm0.20}$ & $86.11_{\pm0.54}$ \\
& FlexMatch~\cite{zhang2021flexmatch} & $87.88_{\pm0.02}$ & $87.92_{\pm0.00}$ & $74.48_{\pm1.18}$ & $77.01_{\pm0.06}$ & $71.79_{\pm0.76}$ & $72.54_{\pm0.84}$ & $82.76_{\pm0.59}$ \\
& SimMatch~\cite{zheng2022simmatch} & $89.32_{\pm0.54}$ & $89.54_{\pm0.45}$ & $75.14_{\pm0.01}$ & $76.98_{\pm0.21}$ & $72.03_{\pm0.78}$ & $73.36_{\pm0.06}$ & $\underline{87.04}_{\pm0.27}$ \\ \midrule
\multirow{3}{*}{\rotatebox[origin=c]{90}{\textcolor{gray}{OSSL}}} & MTC~\cite{yu2020mtc} & $79.14_{\pm9.78}$ & $89.55_{\pm3.20}$ & $66.32_{\pm1.25}$ & $65.40_{\pm1.99}$ & $65.71_{\pm0.09}$ & $66.30_{\pm5.49}$ & $78.61_{\pm1.86}$ \\
 & OpenMatch~\cite{saito2021openmatch} & $44.18_{\pm5.05}$ & $92.35_{\pm1.01}$ & $75.43_{\pm0.11}$ & $78.66_{\pm0.70}$ & $61.78_{\pm5.58}$ & $73.70_{\pm0.04}$ & $76.63_{\pm0.54}$ \\
 & T2T~\cite{huang2021t2t} & $54.68_{\pm7.01}$ & $55.61_{\pm9.43}$ & $67.26_{\pm6.35}$ & $64.44_{\pm0.50}$ & $53.45_{\pm0.27}$ & $69.11_{\pm6.36}$ & $83.88_{\pm0.17}$ \\ \midrule
 & SSB (FixMatch) & $\underline{92.37}_{\pm2.36}$ & $\textbf{95.67}_{\pm0.23}$ & $76.44_{\pm0.15}$ & $\underline{81.16}_{\pm0.61}$ & $\underline{73.18}_{\pm0.38}$ & $\underline{76.34}_{\pm0.14}$ & $81.77_{\pm0.04}$  \\ 
 & SSB (FlexMatch) & $83.83_{\pm1.25}$ & $92.91_{\pm1.03}$ & $\underline{77.40}_{\pm0.82}$ & $80.72_{\pm0.65}$ & $71.38_{\pm0.53}$ & $75.96_{\pm0.99}$ & $83.26_{\pm1.02}$ \\
 & SSB (SimMatch) & $\textbf{95.47}_{\pm0.26}$ & $\underline{95.07}_{\pm0.10}$ & $\textbf{79.80}_{\pm0.13}$ & $\textbf{81.82}_{\pm0.23}$ & $\textbf{76.14}_{\pm2.08}$ & $\textbf{79.27}_{\pm0.72}$ & $\textbf{88.46}_{\pm0.10}$ \\ \bottomrule
\end{tabular}
\end{center}
\caption{
\textbf{AUROC (\%) for unseen outliers. }
The best number is in \textbf{bold}, and the second best is in \underline{underline}.
Note that a random OOD detector gives an AUROC of 50\%.
}
\label{tab:unseen_auroc}
\end{table*}

\section{Results on More Benchmarks} \label{sec:benchmark}
In this section, we compare SSB with more recent methods on their benchmarks.

\myparagraph{Comparison with methods of class-mismatched SSL.}
Here, we compare with Safe-Student \cite{he2022safe} and SPL \cite{he2022not} on CIFAR-10 and CIFAR-100 with different levels of class distribution mismatch between the labeled and unlabeled data.
Following \cite{he2022not,he2022safe}, on CIFAR-10, we consider the six animal classes as inlier classes and use 400 labels per class.
The unlabeled set contains 20,000 images coming from all ten classes with different class mismatch ratios.
For example, when the ratio is 0.3, 70\% of the samples are from the six inlier classes and the rest samples are from the remaining four classes. 
Similarly, for CIFAR-100, the first 50 classes are used as inlier classes, and the unlabeled set has 20,000 samples with different class mismatch ratios.
We compare the inlier accuracy of different methods in Table \ref{tab:safe_student}.
SSB outperforms other methods by significant margins across all settings, which indicates the effectiveness of our method for class-mismatched SSL.

\myparagraph{Comparison in cross-dataset scenarios.}
Now, we compare our method in cross-dataset settings, where the labeled set and the unlabeled set are constructed using different datasets.
Following \cite{huang2022tmm}, we use CIFAR-100 for labeled set and ImageNet for unlabeled set.
Specifically, we take 60 classes of CIFAR-100 as inlier classes, which are also contained in ImageNet.
Then, 20,000 images are sampled from 100 classes of ImageNet to form the unlabeled set, where 60 classes are the same as the inlier classes of CIFAR-100 and the rest 40 are randomly chosen from the remaining 940 classes.
Please refer to \cite{huang2022tmm} for details of the 60 inlier classes.
In Table \ref{tab:tmm}, we compare the results of our method with others under different numbers of labeled data.
We can see that our method improves the SOTA in all settings.
The large performance gap over TOOR \cite{huang2022tmm} shows that the simple confidence-based pseudo-labeling used in SSB is a more effective way for recycling OOD data to improve the classification performance.

\begin{table*}[ht]
\begin{center}
\begin{tabular}{lcccc}
\toprule
\multirow{2}{*}{Method} & \multicolumn{2}{c}{CIFAR-10} & \multicolumn{2}{c}{CIFAR-100} \\ \cmidrule(l){2-3} \cmidrule(l){4-5} 
 & ratio=0.3 & ratio=0.6 & ratio=0.3 & ratio=0.6 \\ \midrule
DS$^3$L \cite{guo2020ds3l} & $78.1_{\pm0.4}$ & $76.9_{\pm0.5}$ & - & - \\
UASD \cite{chen2020uasd} & $77.6_{\pm0.4}$ & $76.0_{\pm0.4}$ & $61.8_{\pm0.4}$ & $58.4_{\pm0.5}$ \\
MTC \cite{yu2020mtc} & $85.5_{\pm0.6}$ & $81.7_{\pm0.5}$ & $63.1_{\pm0.6}$ & $61.1_{\pm0.3}$ \\
CL \cite{cascante2021cl} & $83.2_{\pm0.4}$ & $82.1_{\pm0.4}$ & $63.6_{\pm0.4}$ & $61.5_{\pm0.5}$ \\
Safe-Student \cite{he2022safe} & $85.7_{\pm0.3}$ & $83.8_{\pm0.1}$ & $68.4_{\pm0.2}$ & $68.2_{\pm0.1}$\\
CL+SPL \cite{he2022not} & $87.8_{\pm0.3}$ & $84.1_{\pm0.5}$ & $65.9_{\pm0.3}$ & $65.5_{\pm0.4}$ \\ \midrule
SSB (Ours) & $\textbf{92.5}_{\pm0.1}$ & $\textbf{90.6}_{\pm0.3}$ & $\textbf{74.7}_{\pm0.6}$ & $\textbf{73.2}_{\pm0.3}$ \\ \bottomrule
\end{tabular}
\end{center}
\caption{\textbf{Test accuracy (\%) with different class mismatch ratios on CIFAR-10 and CIFAR-100.} 
This benchmark is adopted by \cite{he2022safe, he2022not}.
The number of unlabeled data is 20,000. The best number is in \textbf{bold}.}
\label{tab:safe_student}
\end{table*}

\begin{table*}[ht]
\begin{center}
\begin{tabular}{lcccc}
	\toprule
    \multirow{2}{*}{Method} & \multicolumn{4}{c}{CIFAR100+ImageNet with different numbers of labeled data} \\ \cmidrule(l){2-5}
	& 4800 (80 for each class)  & 6000 (100 for each class) & 7200 (120 for each class) & 8400 (140 for each class) \\ 
	\midrule[0.6pt]
    UASD~\cite{chen2020uasd} &$42.07_{\pm0.58}$	&$44.90_{\pm0.47}$ &$46.38_{\pm0.79}$	&$48.20_{\pm0.40}$\\
	DS$^3$L~\cite{guo2020ds3l} &$43.99_{\pm0.54}$	&$45.10_{\pm1.25}$	&$47.11_{\pm0.73}$	&$48.96_{\pm0.63}$\\
	MTC~\cite{yu2020mtc} &$45.69_{\pm0.74}$	&$46.34_{\pm0.81}$	&$48.92_{\pm0.58}$	&$50.05_{\pm0.77}$\\
	TOOR~\cite{huang2022tmm} &${47.19}_{\pm0.83}$	&${49.15}_{\pm0.76}$	&${51.34}_{\pm0.65}$	&${52.98}_{\pm0.79}$ \\ \midrule
	SSB (Ours) & $\textbf{63.91}_{\pm0.71}$ & $\textbf{65.95}_{\pm0.20}$ & $\textbf{67.97}_{\pm0.12}$ & $\textbf{69.42}_{\pm0.28}$ \\
	\bottomrule
\end{tabular}
\end{center}
\caption{\textbf{Test accuracy (\%) with different numbers of labeled samples.} 
This benchmark is adopted by \cite{huang2022tmm}.
The number of unlabeled data is 20,000. The best number is in \textbf{bold}.}
\label{tab:tmm}
\end{table*}

\section{Ablation Study} \label{sec:ablation_appendix}
In this section, we provide more analysis of the hyper-parameters used in SSB.
Specifically, we study the effect of depth and width of the projection head, deferred outlier detector training, the effect of the threshold $\theta$ for selecting the pseudo-outliers, the effect of the threshold $\tau$ for pseudo-labeling, the loss weight $\lambda_{det}^u$ for unlabeled detection loss, and different data augmentation schemes used in pseudo-negative mining.

\myparagraph{Effect of depth and width of the projection head.}
Table \ref{tab:proj_depth} and \ref{tab:proj_width} study the effect of the depth and the width of the projection head, respectively. 
We choose the 2-layer MLP with a hidden dimension of 1024 as it shows the best inlier accuracy and OOD detection performance.
Note that the projection head shows good robustness over a large range of design choices.
Even using a single fully-connected layer with a ReLU activation function already gives better performance.

\begin{table}[ht]
\begin{center}
\resizebox{\linewidth}{!}{%
\begin{tabular}{lccc}
\toprule
Projection head & \begin{tabular}[c]{@{}c@{}}Inlier Cls.\\ (Acc.)\end{tabular}  & \begin{tabular}[c]{@{}c@{}}Outlier Det.\\ (seen AUROC)\end{tabular} & \begin{tabular}[c]{@{}c@{}}Outlier Det.\\ (unseen AUROC)\end{tabular} \\ \midrule
None & 90.28 & 44.11 & 82.81 \\ 
1-layer MLP & 91.48 & 98.89 & 89.96 \\
2-layer MLP & \textbf{91.65} & \textbf{99.16} & \textbf{90.35} \\
3-layer MLP & 91.48 & 98.49 & 88.08 \\
4-layer MLP & 91.28 & 98.29 & 86.44 \\
2-layer MLP + BN & 90.00 & 90.53 & 78.33 \\ \bottomrule
\end{tabular}
}%
\end{center}
\caption{
\textbf{Effect of different projection heads.}
\textit{None} denotes not using projection head; \textit{MLP} denotes multilayer perceptron; \textit{BN} denotes using BatchNorm \cite{ioffe2015batchnorm} between different layers.
All models are trained with confidence-based pseudo-labeling and pseudo-negative mining on the same data split on CIFAR-10 with 25 labeled data.
}
\label{tab:proj_depth}
\end{table}



\begin{table}[ht]
\begin{center}
\resizebox{\linewidth}{!}{%
\begin{tabular}{lccc}
\toprule
Hidden dim. & \begin{tabular}[c]{@{}c@{}}Inlier Cls.\\ (Acc.)\end{tabular}  & \begin{tabular}[c]{@{}c@{}}Outlier Det.\\ (seen AUROC)\end{tabular} & \begin{tabular}[c]{@{}c@{}}Outlier Det.\\ (unseen AUROC)\end{tabular} \\ \midrule
128 & 91.25 & 97.30 & 82.98 \\
256 & 90.83 & 97.47 & 85.43 \\
512 & 90.63 & 98.91 & 88.36 \\
1024 & \textbf{91.65} & \textbf{99.16} & \textbf{90.35} \\
2048 & 91.07 & 97.74 & 86.59 \\ \bottomrule
\end{tabular}
}%
\end{center}
\caption{
\textbf{Effect of the hidden dimension of the projection head.}
We use a 2-layer MLP as the projection and train all models with confidence-based pseudo-labeling and pseudo-negative mining on the same data split on CIFAR-10 with 25 labeled data.
}
\label{tab:proj_width}
\end{table}

\myparagraph{Deferring the outlier detector training.}
Here we study the effect of different training lengths on the outlier detector.
In Table \ref{tab:defer}, we can see that while the performance are similar under different training epochs, the training cost can be largely reduced by deferring the detector training.
With 37 training epochs, our method can reach the best performance while reducing the training costs. 

\begin{table}[ht]
\begin{center}
\resizebox{\linewidth}{!}{%
\begin{tabular}{cccccc}
\toprule
\begin{tabular}[c]{@{}c@{}}Total\\ epochs\end{tabular} & \begin{tabular}[c]{@{}c@{}}Starting\\ epochs\end{tabular} & \begin{tabular}[c]{@{}c@{}}GPU hours\\ reduced\end{tabular} & \begin{tabular}[c]{@{}c@{}}Inlier Cls.\\ (Acc.)\end{tabular}  & \begin{tabular}[c]{@{}c@{}}Outlier Det.\\ (avg. AUROC)\end{tabular}\\ \midrule
512 & 0 & 0 & 90.20 & 87.92 \\
512 & 100 & 5.9 & 90.58 & 89.11 \\
512 & 200 & 10.7 & 91.01 & 90.39 \\
512 & 300 & 16.0 & 91.55 & 91.71 \\
512 & 400 & 19.6 & 91.62 & 93.13 \\
512 & 475 & \textbf{23.2} & \textbf{91.65} & \textbf{94.76} \\ \bottomrule
\end{tabular}
}%
\end{center}
\caption{
\textbf{Effect of different starting epochs for outlier detector training.}
We defer the detector training by enabling the detection losses at a later stage.
All models are trained on a single NVIDIA V100 and we compute the reduced GPU hours with respect to the model using detection losses throughout the entire training.
The setting is CIFAR-10 with 25 labeled data.
}
\label{tab:defer}
\end{table}



\myparagraph{Effect of threshold $\tau$.} We follow FixMatch \cite{sohn2020fixmatch} and set the threshold $\tau$ for pseudo-labeling as 0.95 for our main results.
Here we provide an ablation study of this hyper-parameter in Table \ref{tab:tau}, with SSB exhibiting robustness within a wide range of $\tau$.

\begin{table}[ht]
\begin{center}
\resizebox{0.9\linewidth}{!}{%
\begin{tabular}{lccccc}
\toprule
Threshold $\tau$    & 0.97 & 0.95           & 0.85 & 0.75 & 0.5 \\ \midrule
Inlier Cls.  &    90.28  & 91.65 &   92.28   &   90.52   &  \textbf{92.35}   \\
Outlier Det. &   93.67   & 94.76 &  \textbf{97.41}    &   97.17   &   92.64  \\ \bottomrule
\end{tabular}
}%
\end{center}
\caption{\textbf{Effect of different pseudo-labeling thresholds $\tau$}. SSB shows good robustness within a wide range of $\tau$. The experimental setting is CIAR-10 with 25 labels.}
\label{tab:tau}
\end{table}

\myparagraph{Different loss weights.}
In the main paper, we use $\lambda_{det}^u=1$ for the unlabeled data detection loss with pseudo-negative mining.
Here, we provide more results of different loss weights in Table \ref{tab:pseudo_neg_lambda}.
We can see that, except for very small loss weights (0.1 or 0), the OOD detection performance is quite robust to various $\lambda_{det}^u$.

\begin{table}[ht]
\begin{center}
\resizebox{\linewidth}{!}{%
\begin{tabular}{cccc}
\toprule
\begin{tabular}[c]{@{}c@{}}Loss weight\\ $\lambda_{det}^u$\end{tabular} & \begin{tabular}[c]{@{}c@{}}Inlier Cls.\\ (Acc.)\end{tabular}  & \begin{tabular}[c]{@{}c@{}}Outlier Det.\\ (seen AUROC)\end{tabular} & \begin{tabular}[c]{@{}c@{}}Outlier Det.\\ (unseen AUROC)\end{tabular} \\ \midrule
10 & \textbf{92.45} & 99.18 & 89.40 \\
5 & 91.98 & 98.61 & 88.20 \\
2 & 91.98 & 98.82 & 88.96 \\
1 & 91.65 & 99.16 & 90.35 \\
0.5 & 92.08 & \textbf{99.22} & 90.18 \\
0.1 & 91.77 & 88.98 & \textbf{90.74} \\
0 & 91.52 & 89.78 & 90.27 \\
\bottomrule
\end{tabular}
}%
\end{center}
\caption{
\textbf{Effect of the loss weight for pseudo-negative mining.}
Our method is robust to a wide range of loss weights.
The setting is CIFAR-10 with 25 labeled data.
}
\label{tab:pseudo_neg_lambda}
\end{table}

\myparagraph{Effect of data augmentation scheme.} In pseudo-negative mining, there are two types of data augmentations used for loss computing.
Given an unlabeled image, the OOD score is computed from a weak augmentation, which consists of random crop and horizontal flipping.
Then, if the confidence is low enough, a strong augmentation will be used to compute the binary cross-entropy loss.
Following \cite{sohn2020fixmatch}, the strong augmentation is RandAugment \cite{cubuk2020randaugment} with CutOut \cite{devries2017cutout}.
In Table \ref{tab:pseudo_neg_DA}, we study the effect of different types of augmentation for computing the loss.
Using the strong augmentation gives the best OOD detection performance while having similar inlier classification performance to other strategies.

\begin{table}[ht]
\begin{center}
\resizebox{\linewidth}{!}{%
\begin{tabular}{cccc}
\toprule
Augmentation & \begin{tabular}[c]{@{}c@{}}Inlier Cls.\\ (Acc.)\end{tabular}  & \begin{tabular}[c]{@{}c@{}}Outlier Det.\\ (seen AUROC)\end{tabular} & \begin{tabular}[c]{@{}c@{}}Outlier Det.\\ (unseen AUROC)\end{tabular} \\ \midrule
weak+strong & \textbf{91.88} & 92.36 & 77.55 \\
weak & 91.73 & 92.10 & 77.80 \\
strong & 91.65 & \textbf{99.16} & \textbf{90.35} \\
\bottomrule
\end{tabular}
}%
\end{center}
\caption{
\textbf{Effect of data augmentation in pseudo-negative mining. }
It is important to use a different type of data augmentation for loss computing from the one used to generate pseudo-outliers. 
The setting is CIFAR-10 with 25 labeled data.
}
\label{tab:pseudo_neg_DA}
\end{table}


\section{Pseudo-Code} \label{sec:pseudo_code}
We present the complete pseudo-code of SSB with deferred outlier detector training in Algorithm \ref{alg1}.

\section{Visualization of Pseudo-Inliers} \label{sec:visualization}
\begin{figure*}[t]
\centering
\includegraphics[width=1\linewidth]{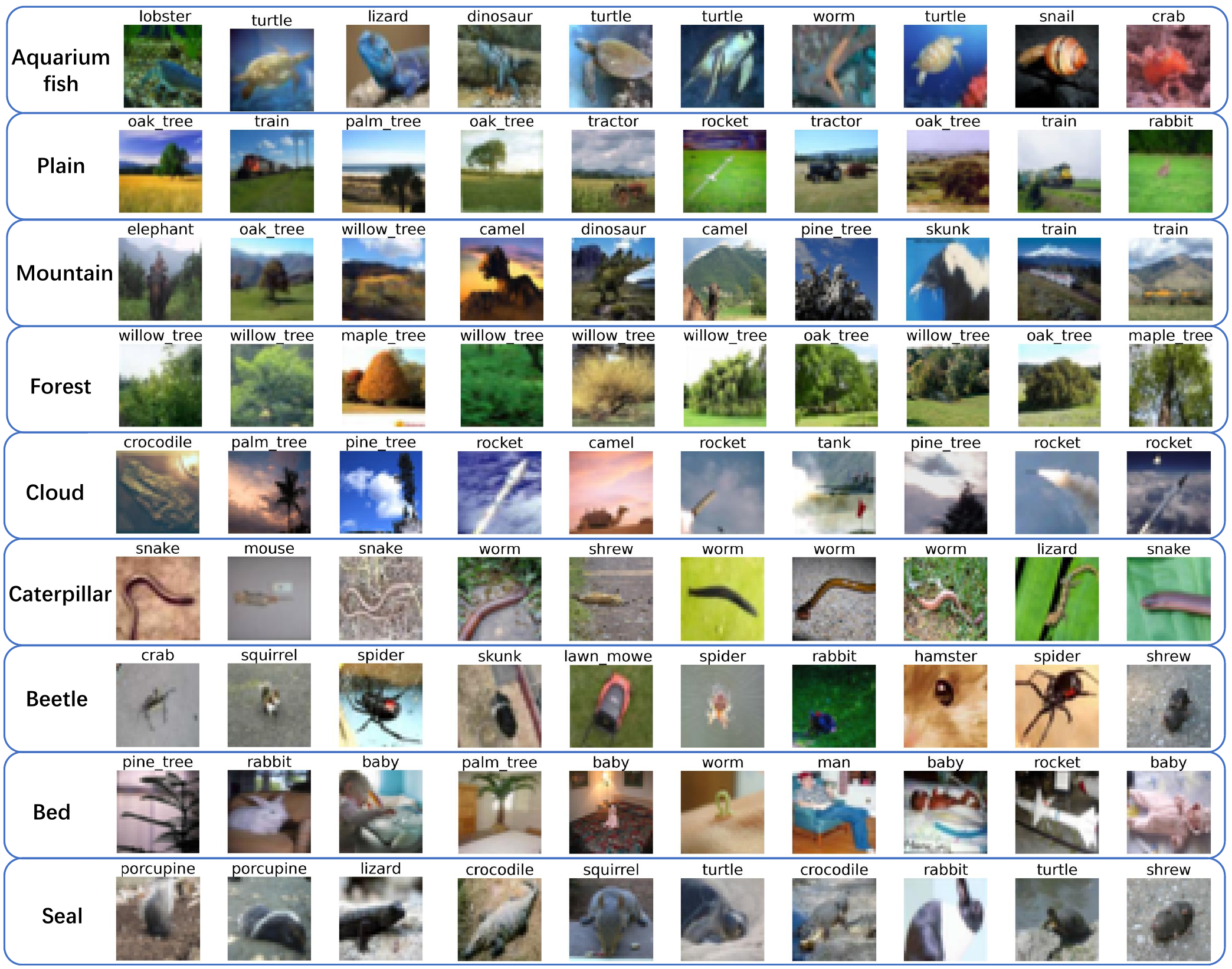}
\caption{
Selected pseudo-inliers for different classes.
Each row lists 10 most confident images with the pseudo-label on the left of the row.
The ground-truth class of the OOD sample is shown on the top of each image.
}
\label{fig:appendix_ood}
\end{figure*}

We visualize the OOD samples selected for different classes in Fig. \ref{fig:appendix_ood}.
The model is trained on CIFAR-100 with 55 inlier classes and 25 labels.
We can see that the selected pseudo-inliers contain semantic information of the corresponding classes, which indicates that some OOD data are natural data augmentations and can be used to improve the generalization performance if used properly.

\begin{algorithm*}[ht]
\caption{}
\begin{algorithmic}[1]  \label{alg1}
    \STATE {\bfseries Input:} Labeled set $\mathcal{D}_{\text{labeled}}=\{(\textbf{x}_i^l, y_i)\}_{i=1}^N$, unlabeled set $\mathcal{D}_{\text{unlabeled}}=\{(\textbf{x}_i^u)\}_{i=1}^M$, feature encoder $f$, inlier classifier $g_c$, outlier detector $g_d$, two MLP projection heads $h_c$ and $h_d$, thresholds $\tau$ and $\theta$, batch size $B_l$ and $B_u$, loss weights $\lambda_{det}^u$, $\lambda_{OC}^u$, and $\lambda_{em}^u$, warm-up iterations $T_0$, total number of training iterations $T$ \\
    \STATE Initialize the parameters of $f$, $g_c$, $g_d$, $h_c$, and $h_d$ randomly
    %
    \FOR{$t = 1$ \TO $T$}
    \STATE \COMMENT{Sample labeled and unlabeled data} \\
    \STATE $\{\textbf{x}_i^l, y_i\}_{i=1}^{B_l}$ $\sim$ Random sampler($\mathcal{D}_{\text{labeled}}$) \\
    \STATE $\{\textbf{x}_i^u\}_{i=1}^{B_u}$ $\sim$ Random sampler($\mathcal{D}_{\text{unlabeled}}$) \\
    \STATE \COMMENT{Compute classification losses} \\
    \STATE $\hat{p}_i^u = \text{softmax}(g_c(h_{c}(f(\textbf{x}_i^u))), i=1,...,B_u$ \COMMENT{Compute pseudo-label distributions}
    \STATE $\hat{y}^u_i = \text{argmax} \hat{p}_i^u, i=1,...,B_u$ \COMMENT{Compute pseudo-labels}
    \STATE $L_{cls}^l = \frac{1}{B_l} \sum_{i=1}^{B_l} H(g_c(h_{c}(f(\textbf{x}_i^l)), y_i)$ \COMMENT{Labeled data loss}
    \STATE $L_{cls}^u = \frac{1}{B_u} \sum_{i=1}^{B_u} \mathbbm{1}(\max \hat{p}_i^u \geq \tau) H(\hat{p}_i^u, \hat{y}^u_i)$ \COMMENT{Unlabeled data loss as in Equation (2)}
    \STATE $L_{cls} = L_{cls}^l + L_{cls}^u$ 
    \STATE \COMMENT{Compute detection losses} \\
    \STATE $L_{det}^l = - \frac{1}{B_l} \sum_{i=1}^{B_l} log(p_{y_i}(\textbf{x}_i^l)) + \frac{1}{|\mathcal{C}|-1} \sum_{j\neq y_i} log(1 - p_{j}(\textbf{x}_i^l))$ \COMMENT{Detection loss for labeled data as in Equation (4)} \\
    \STATE $L_{det}^u = - \frac{1}{B_u} \sum_{i=1}^{B_u} \frac{1}{\sum_c{\mathbbm{1}(p_c > \theta)}} \sum_{c=1}^{|\mathcal{C}|} \mathbbm{1}(p_c > \theta) log(1 - p_{c}(\textbf{x}_i^u))$ \COMMENT{Pseudo-negative mining as in Equation (5)} \\
    \STATE $L_{em}^{u} = \frac{1}{B_u} \sum_{i=1}^{B_u} entropy(g_d(h_d(f(\textbf{x}_i^u)))) $ \COMMENT{Entropy minimization loss as in \cite{grandvalet2005minentropy}} \\
    \STATE $L_{OC}^{u} = \frac{1}{B_u} \sum_{i=1}^{B_u} ||g_d(h_d(f(\mathcal{T}_1(\textbf{x}_i^u)))) - g_d(h_d(f(\mathcal{T}_2(\textbf{x}_i^u))))||^2 $ \COMMENT{Open-set consistency loss as in \cite{saito2021openmatch}} \\
    \STATE  $L_{det} = L_{det}^l + \lambda_{det}^u L_{det}^u + \lambda_{OC}^u L_{OC}^{u} + \lambda_{em}^u L_{em}^{u}$
    \STATE \COMMENT{Total loss}
    \STATE $L_{total} = L_{cls} + \mathbbm{1}(t>T_0)L_{det}$
    \STATE Update parameters in $f$, $g_c$, $g_d$, $h_c$, and $h_d$ with SGD
    \ENDFOR
    \RETURN $f$, $g_c$, $g_d$, $h_c$, and $h_d$ \\
\end{algorithmic}
\end{algorithm*}

\end{document}